\documentclass{article} 
\usepackage{iclr2026_conference,times}

\usepackage{microtype}
\usepackage{graphicx}
\usepackage{subcaption}
\usepackage{booktabs,multirow,makecell}
\usepackage{placeins}

\usepackage{amsmath}
\usepackage{amssymb}
\usepackage{mathtools}
\usepackage{amsthm}

\usepackage{hyperref}
\usepackage{url}
\usepackage[capitalize,noabbrev]{cleveref}

\usepackage{pifont}
\usepackage{stfloats}

\newcommand{\cmark}{\ding{51}}
\newcommand{\xmark}{\ding{55}}

\usepackage{algorithm}
\usepackage{algorithmic}
\usepackage[most]{tcolorbox}
\usepackage{longtable}
\usepackage{wrapfig}
\usepackage{mdframed}
\usepackage{enumitem}
\usepackage{xcolor}
\definecolor{ForestGreen}{RGB}{34,139,34}

\newtcolorbox{promptbox}[1]{%
  enhanced,
  breakable,
  colback=gray!5,
  colframe=black!30,
  boxrule=0.5pt,
  arc=2pt,
  left=6pt,right=6pt,top=6pt,bottom=6pt,
  fonttitle=\bfseries,
  title={#1}
}
\usepackage{colortbl}

\tcbuselibrary{listingsutf8}

\newtcblisting{respgenprompt}[2][]{%
  enhanced, breakable,
  colback=black!2, colframe=blue!55!black,
  title={#2}, fonttitle=\bfseries, 
  colbacktitle=blue!55!black, 
  coltitle=white,              
  listing only,
  listing options={basicstyle=\small\ttfamily, breaklines=true},
  #1
}

\newtcblisting{verifierprompt}[2][]{%
  enhanced, breakable,
  colback=orange!3,
  colframe=orange!70!black,
  title={#2},
  fonttitle=\bfseries,
  colbacktitle=orange!70!black,
  listing only,
  listing options={basicstyle=\small\ttfamily, breaklines=true},
  #1
}
\newtcblisting{judgeprompt}[2][]{%
  enhanced, breakable,
  colback=violet!3,
  colframe=violet!65!black,
  title={#2},
  fonttitle=\bfseries,
  colbacktitle=violet!65!black,
  listing only,
  listing options={basicstyle=\small\ttfamily, breaklines=true},
  #1
}

\title{Quantifying Consistency in LLM Logical Reasoning via Structural Uncertainty}

\iclrfinalcopy
\author{
Baishali Chaudhury, Mengdie Flora Wang, Hyunji Hayley Park, Rahul Ghosh, \\ 
\textbf{ Sungmin Hong, Jae Oh Woo}\\
 { AWS Generative AI Innovation Center } \\
\small\texttt{\{baishch, florawan, parhyunj, rahulgh, hsungmin, jaeohwoo\}@amazon.com}
}

\begin{document}
\maketitle

\begin{abstract}
Large language models can arrive at the same answer through reasoning paths that are unstable, contradictory, or difficult to rank consistently---a failure mode especially prevalent in multi-step deductive reasoning. Existing methods assess reasoning reliability primarily through output dispersion---measuring how much sampled answers differ---but this view discards a complementary signal: whether the model can consistently rank competing reasoning candidates. We propose structural uncertainty, a consistency-aware evaluation framework derived from the stability of self-preference-induced rankings over sampled reasoning solutions. Given a query, we generate multiple candidate solutions and ask the same model to judge pairwise preferences among its own outputs. We aggregate sparse self-preferences into ranking distributions via Bradley--Terry modeling with PageRank, and decompose the signal into two complementary entropy-based components---across-trial ranking instability and within-trial candidate ambiguity. Across five LLMs and eight benchmarks, structural signals provide information complementary to answer dispersion: on logical and mathematical reasoning tasks, the combination improves identification of unreliable reasoning instances, while on factual retrieval the structural signal collapses toward uniformity, diagnosing a regime boundary where reasoning-level consistency evaluation is uninformative. The two components relate differently to accuracy: within-trial ambiguity correlates positively with correctness on reasoning tasks---consistent with settings where multiple plausible solution paths remain competitive---while across-trial instability correlates negatively, signaling unreliable reasoning. Structural uncertainty is best understood not as a universal confidence estimator, but as a regime-sensitive evaluator of logical reasoning consistency.
\end{abstract}

\section{Introduction}
\begin{figure}[!t]
  \centering
  \includegraphics[width=0.65\columnwidth]{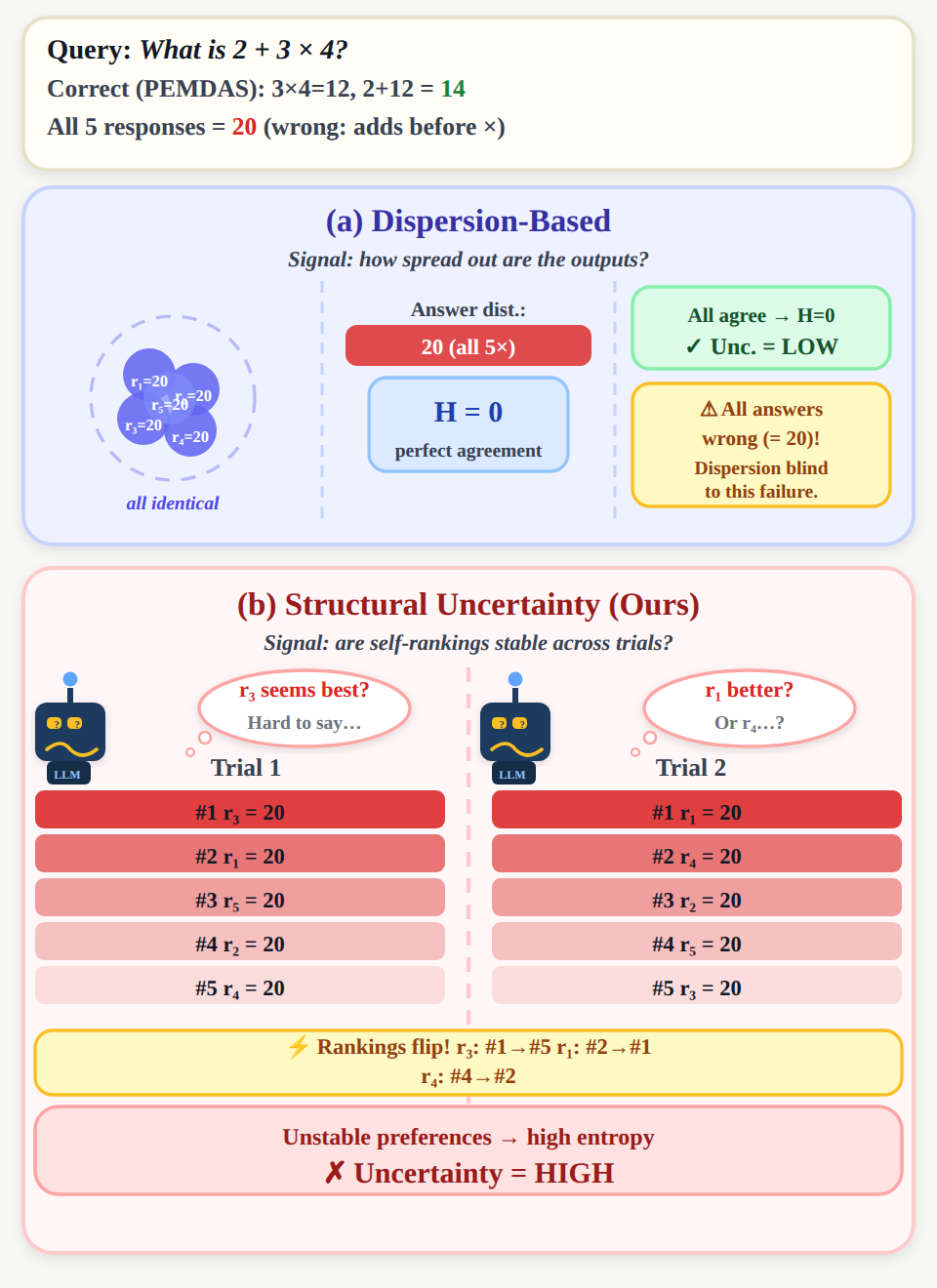}
  \caption{An illustrative example contrasting dispersion-based and structural
  consistency evaluation. \textbf{(a)} When all five sampled reasoning candidates agree on the same
  wrong answer (20), dispersion-based methods report zero entropy and low
  uncertainty, since they do not use preference-order information among candidates. \textbf{(b)} Our structural
  approach asks the model to rank its own reasoning outputs across independent trials.
  The rankings are completely unstable (r\textsubscript{3} moves from \#1 to
  \#5), revealing reasoning inconsistency not reflected by answer dispersion alone.}
  \label{fig:dispersion_vs_structural}
\end{figure}
Large language models have achieved remarkable progress in natural language understanding and generation, yet their logical reasoning capabilities remain a significant bottleneck \cite{xiong2023can, tian2023just, kapoor2024calibration, zhou2024relying}. Models frequently produce answers that appear logically plausible yet are internally inconsistent---arriving at the same wrong conclusion through different flawed reasoning paths, or failing to maintain stable preferences when asked to compare their own solutions. Evaluating logical reasoning reliability requires assessing not only answer-level correctness but also the \emph{consistency} of the reasoning process itself. Although we do not directly model cross-question contradiction, our framework targets a closely related consistency problem: whether the model can stably evaluate competing reasoning candidates for the same query.

Existing post-hoc evaluation methods \citep{lin2023generating, farquhar2024detecting, wang2024self, lyu2025calibrating} treat sampled responses as exchangeable and assess reliability from how much answers \emph{differ}---i.e., output dispersion. This view captures one dimension of reasoning reliability but discards a complementary signal: the \emph{structural consistency} of preferences among candidate reasoning solutions. For logical reasoning, this omission is consequential because multiple candidates may share the same final answer yet differ in reasoning quality, coherence, or mutual consistency; collapsing this structure loses information about which solutions the model favors and how stable those preferences are. Figure~\ref{fig:dispersion_vs_structural} illustrates this gap: when all sampled responses agree on the same wrong answer, dispersion-based methods report low uncertainty, while self-preference rankings across trials can reveal instability not reflected by dispersion alone.

We propose a consistency-aware evaluation framework for logical reasoning that measures the stability of self-preference-induced rankings over sampled candidate solutions. Given a query, we sample multiple reasoning candidates and ask the \emph{same model} to judge pairwise preferences among its own outputs. Beyond how much responses \emph{differ}, this probes whether the model forms a stable or fluctuating preference ordering over competing reasoning paths---a signal that dispersion alone discards. We aggregate sparse pairwise judgments into ranking distributions using Bradley--Terry modeling \citep{BradleyTerry1952} with PageRank normalization \citep{LangvilleMeyer2006PageRank}, repeated across random spanning-tree comparison trials. This produces two distinct components of structural uncertainty: \emph{across-trial ranking instability} (reasoning instability across elicitation trials) and \emph{within-trial candidate ambiguity} (ambiguity among plausible reasoning candidates within each trial).

Across five LLMs and eight benchmarks, we show that self-preference-derived structural signals provide information complementary to output dispersion. The interaction with accuracy is task-dependent: on mathematical reasoning, within-trial ambiguity can correlate positively with correctness---consistent with settings where several plausible solution paths remain competitive---while on factual retrieval, structural signals collapse toward uniformity, diagnosing a regime where reasoning-level consistency evaluation is uninformative. Combining structural and dispersion-based signals improves identification of unreliable reasoning instances on several reasoning and knowledge tasks, though gains are absent in the collapse regime.

\paragraph{Contributions.}
\textbf{(1) A post-hoc framework for evaluating logical reasoning consistency and reliability.} We propose a model-agnostic framework that quantifies reasoning consistency through the stability of a model's self-preference rankings over its own candidate solutions, providing an observable signal complementary to output dispersion.
\noindent\textbf{(2) A structural decomposition of reasoning stability.} We aggregate sparse pairwise self-preferences into ranking distributions via Bradley--Terry with PageRank, yielding two complementary entropy-based components: across-trial ranking instability (reasoning instability) and within-trial candidate ambiguity (ambiguity among plausible reasoning candidates).
\noindent\textbf{(3) A regime analysis distinguishing reasoning from retrieval settings.} Across five LLMs and eight benchmarks, we show that structural signals complement dispersion on reasoning tasks, identify when and why the signal collapses on retrieval tasks, and characterize the task conditions under which each signal type is most informative for evaluating logical reasoning consistency.

\section{Related Work}
\label{sec:related}

\paragraph{Logical reasoning and self-consistency in LLMs.}
Chain-of-thought prompting and self-consistency \citep{wang2023selfconsistency} have become standard approaches for improving and evaluating LLM reasoning. Self-consistency measures agreement across multiple sampled reasoning paths, while debate and self-judge frameworks \citep{zheng2023judging, kadavath2022language} demonstrate that models can assess output quality. However, self-consistency treats responses as exchangeable and measures only answer-level agreement, missing structural differences in reasoning quality among candidates. We complement answer-level consistency by measuring how stably the model ranks competing reasoning solutions through self-preference.

\paragraph{Post-hoc uncertainty and preference-based evaluation.}
Dispersion-based methods estimate uncertainty from semantic variation among responses \citep{kuhn2023semantic, lin2023generating, farquhar2024detecting, kossen2024semantic}, output density \citep{qiu2024semantic}, kernelized entropy \citep{nikitin2024kernel}, or self-consistency entropy \citep{wang2024self, lyu2025calibrating}. Comparison-based methods aggregate pairwise preferences into calibrated scores \citep{shrivastava2025language} or incorporate richer structure via multi-dimensional representations \citep{chen2025uncertainty}, knowledge graphs \citep{yuan2025kg}, or Minimum Bayes Risk \citep{vashurin2025cocoa}. When internal access is available, logit-based \citep{ma2025estimating}, chain-of-thought \citep{zhang2025cot}, and proxy-based methods \citep{lee2024improving} derive uncertainty from model internals. Information-theoretic and Bayesian perspectives motivate principled decomposition \citep{abbasi2024believe, KendallGal2017Uncertainties, woo2022analytic, woo2023active}. Our approach operates in a fully black-box setting without requiring internal access or model modification.

\paragraph{Consistency and contradiction in LLM outputs.}
A growing body of work addresses logical contradictions and inconsistencies in LLM outputs. Evaluation frameworks assess alignment between uncertainty and quality \citep{huang2024uncertainty, ye2024benchmarking, vashurin2025benchmarking}, while studies reveal strong task- and model-dependence in LLM reliability \citep{huang2023look, yang2025maqa}. Our work contributes to this direction by providing a structural lens on reasoning consistency: rather than checking whether outputs contradict each other at the answer level, we measure whether the model can form a stable preference ordering over its own reasoning candidates---directly revealing internal inconsistency in reasoning evaluation. Unlike approaches that improve reasoning via symbolic modules or external solvers, we focus on post-hoc evaluation of reasoning consistency in a pure black-box setting.

\begin{figure*}[t]
\centering
\includegraphics[width=0.9\textwidth]{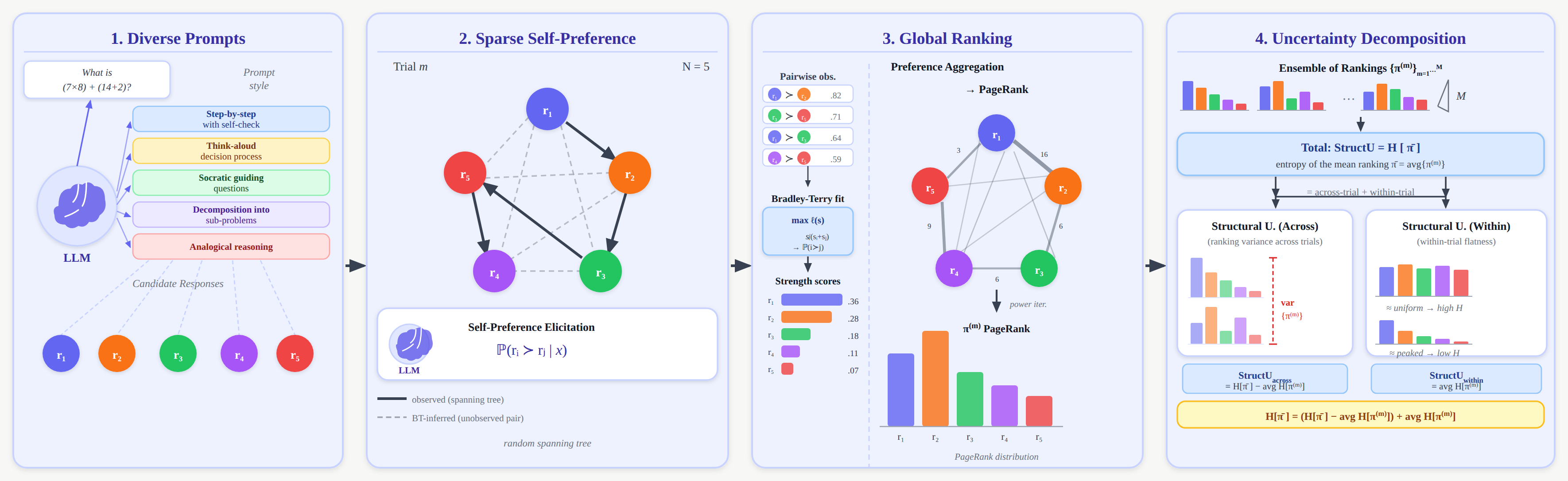}
\caption{Overview of the consistency-aware reasoning evaluation framework.
Given a query, we (1) generate diverse candidate reasoning solutions,
(2) elicit pairwise self-preferences,
(3) aggregate into a global ranking via pairwise preference modeling
(Bradley--Terry or TrueSkill) with PageRank, and
(4) decompose the consistency signal via random spanning tree sampling.}
\label{fig:pipeline}
\end{figure*}

\section{Method}
\label{sec:method}

Our framework quantifies consistency in logical reasoning via the stability of self-preference-induced rankings over sampled candidate solutions. Our notion of consistency is same-query and candidate-relative: we ask whether the model forms a stable preference ordering over multiple reasoning solutions to the same problem. Given a query, we: (1) generate $N$ diverse candidate reasoning solutions, (2) elicit pairwise self-preferences by asking the model to judge its own outputs, (3) aggregate preferences into a global ranking via Bradley--Terry with PageRank, and (4) decompose the consistency signal into across-trial and within-trial components through random spanning tree sampling. Figure~\ref{fig:pipeline} illustrates the pipeline.

\subsection{Self-Preference via Spanning Trees}
\label{sec:self_preference}

Given input $x$, we sample $N$ candidates $\mathcal{R}(x) = \{r_1,\ldots,r_N\}$ from the model's conditional distribution $p_\theta(\cdot|x)$ via diverse prompting with stochastic decoding. Rather than comparing all $\binom{N}{2}$ pairs, we repeat the following for $M$ independent trials ($m = 1,\ldots,M$):

\noindent\textbf{Sparse graph sampling.} We draw a uniform random spanning tree $\mathcal{T}^{(m)}$ over the $N$ candidates, yielding a connected graph with exactly $N{-}1$ edges. This guarantees global connectivity with minimal comparisons while injecting structural randomness across trials, loosely analogous in spirit to Monte Carlo dropout over graph structure rather than weights.

\noindent\textbf{Self-preference elicitation.} For each edge $(i,j) \in \mathcal{T}^{(m)}$, we query the \emph{same model} to judge which response is better, optionally obtaining a confidence score. The consistency of these elicited self-preference judgments across trials provides an additional uncertainty signal that is not directly available from output dispersion alone.

\noindent\textbf{Preference aggregation.} We fit a pairwise preference model (Bradley--Terry or TrueSkill) on the $N{-}1$ observed comparisons to infer win probabilities for \emph{all} $N^2$ pairs, then aggregate into a global ranking distribution $\boldsymbol{\pi}^{(m)}$ via PageRank. Details follow in Section~\ref{sec:aggregation}.

The ensemble $\{\boldsymbol{\pi}^{(m)}\}_{m=1}^M$ from $M$ trials enables the uncertainty decomposition in Section~\ref{sec:uncertainty}.

\subsection{Preference Aggregation}
\label{sec:aggregation}

\noindent\textbf{Bradley--Terry with L2 regularization.}
We assign each candidate $i$ a latent utility $\theta_i \in \mathbb{R}$ and model pairwise preferences as \citep{BradleyTerry1952}:
\begin{equation}
\mathbb{P}(i \succ j) = \frac{\exp(\theta_i)}{\exp(\theta_i) + \exp(\theta_j)}.
\label{eq:bt}
\end{equation}
Since spanning trees admit perfect total orderings, the unregularized maximum likelihood objective is unbounded \citep{Ford1957}. We add an $L^2$-penalty for numerical well-posedness:
\begin{equation}
\mathcal{L}_{\text{reg}}(\boldsymbol{\theta}) = \sum_{(i,j) \in \mathcal{T}^{(m)}} \log \mathbb{P}(i \succ j) - \frac{1}{2C}\|\boldsymbol{\theta}\|^2,
\label{eq:bt_reg}
\end{equation}
where $C > 0$ is the inverse regularization strength. We set $C = 1$ based on an ablation sweep (Appendix~\ref{app:c_sweep}): performance degrades below $C < 1$ but plateaus stably for $C \geq 1$ across all models, confirming bounded parameter estimates. We optimize via MM-style updates \citep{Hunter2004MMBradleyTerry} and evaluate Eq.~\eqref{eq:bt} for all pairs to obtain $\mathbf{P}^{(m)} \in [0,1]^{N \times N}$.

As a robustness check, we also implement a TrueSkill variant \citep{herbrich2006trueskill} with confidence-weighted updates. Despite fundamentally different assumptions, both backends produce highly similar uncertainty rankings (Appendix~\ref{app:ts}), suggesting that the induced ranking distributions are reasonably stable with respect to the choice of preference backend.

\noindent\textbf{PageRank global ranking.}
While BT interpolates sparse comparisons into a dense preference matrix $\mathbf{P}^{(m)}$, PageRank summarizes it into a normalized ranking distribution $\boldsymbol{\pi}^{(m)} \in \Delta^N$ on which we define the entropy-based decomposition (Section~\ref{sec:uncertainty}).
From $\mathbf{P}^{(m)}$, we construct a row-stochastic transition matrix where probability mass flows from weaker to stronger candidates:
\begin{equation}
T^{(m)}_{ij} \propto P^{(m)}_{ji}, \quad T^{(m)}_{ii} = 0,
\end{equation}
with row normalization. We compute the stationary distribution $\boldsymbol{\pi}^{(m)}$ satisfying $\boldsymbol{\pi}^{(m)} = (\mathbf{T}^{(m)})^\top \boldsymbol{\pi}^{(m)}$ via power iteration \citep{BrinPage1998Anatomy,LangvilleMeyer2006PageRank}.

\subsection{Structural Uncertainty Decomposition}
\label{sec:uncertainty}

\begin{figure*}[t]
\centering
\includegraphics[width=0.8\textwidth]{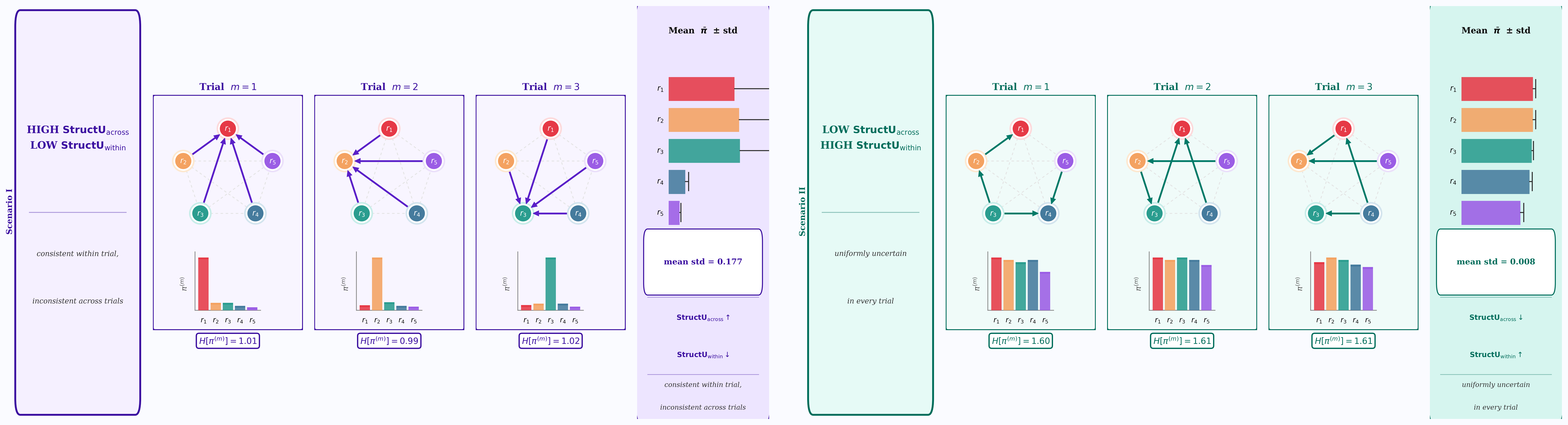}
\caption{\textbf{Across-trial vs.\ within-trial structural uncertainty: two contrasting examples.}
\textbf{Scenario~I (left):} The model produces a confident, concentrated ranking within 
each trial (low within-trial entropy $H[\boldsymbol{\pi}^{(m)}]$), but the dominant 
candidate changes across trials as different spanning trees are sampled---indicating 
substantial across-trial instability in the induced ranking distribution 
($\mathbf{StructU}_{\mathrm{across}}\uparrow$, 
$\mathbf{StructU}_{\mathrm{within}}\downarrow$).
\textbf{Scenario~II (right):} The ranking distribution is nearly uniform in every 
trial (high within-trial entropy), yet remains consistent across trials---indicating 
the model is stably uncertain rather than inconsistently confident 
($\mathbf{StructU}_{\mathrm{across}}\downarrow$, 
$\mathbf{StructU}_{\mathrm{within}}\uparrow$).
Node size reflects PageRank score $\boldsymbol{\pi}^{(m)}$; arrows indicate 
self-preference direction on the sampled spanning tree edges; dashed edges are 
unobserved pairs.}
\label{fig:epistemic_vs_aleatoric}
\end{figure*}

The $M$ trials yield an ensemble $\{\boldsymbol{\pi}^{(m)}\}_{m=1}^M$ with mean $\bar{\boldsymbol{\pi}} = \tfrac{1}{M}\sum_m \boldsymbol{\pi}^{(m)}$. We decompose total uncertainty via Shannon entropy $H[\boldsymbol{p}] = -\sum_i p_i \log p_i$ following \citet{KendallGal2017Uncertainties,woo2022analytic}:
\begin{align}
\text{StructU} &= H[\bar{\boldsymbol{\pi}}], \label{eq:structu_total}\\
\text{StructU}_{\text{within}} &= \tfrac{1}{M}\sum_{m=1}^M H[\boldsymbol{\pi}^{(m)}], \label{eq:structu_within}\\
\text{StructU}_{\text{across}} &= \text{StructU} - \text{StructU}_{\text{within}}. \label{eq:structu_across}
\end{align}
Here, $\mathrm{StructU}_{\text{within}}$ is intended to reflect \emph{within-trial candidate ambiguity}: it is high when, within a single sparse comparison trial, the ranking distribution spreads its mass over multiple candidates instead of concentrating on a single preferred response. Meanwhile, $\mathrm{StructU}_{\text{across}}$ is intended to reflect \emph{across-trial ranking instability}: it is high when different sampled comparison trees lead to substantially different ranking distributions across trials. Figure~\ref{fig:epistemic_vs_aleatoric} illustrates the two components with contrasting examples.

Concretely, $H[\bar{\boldsymbol{\pi}}]$ is the entropy of the trial-averaged ranking; the average $\frac{1}{M}\sum_m H[\boldsymbol{\pi}^{(m)}]$ measures within-trial spread; and their difference---a Jensen gap---provides a distribution-level measure of cross-trial variation. This decomposition is defined entirely over the observable ranking ensemble induced by self-preference; we do not claim it identifies underlying data or parameter uncertainty.

\noindent\textbf{Combination with self-consistency.}
For practical reasoning reliability evaluation, we also study a simple fixed combination of structural uncertainty with self-consistency entropy $\text{Self-ConsU} = -\sum_{a \in \mathcal{A}} p(a) \log p(a)$ over the answer distribution, where $\mathcal{A}$ denotes distinct answers. Abbreviating $\text{StructU}_{\text{across}}$ as $\text{SU}_{\text{a}}$ and $\text{StructU}_{\text{within}}$ as $\text{SU}_{\text{w}}$:
\begin{align}
\text{SU}_{\text{a}}\text{+SC} &= \text{SU}_{\text{a}} + \text{Self-ConsU}, \label{eq:hybrid_across}\\
\text{SU}_{\text{w}}\text{+SC} &= \text{SU}_{\text{w}} - \text{Self-ConsU}, \label{eq:hybrid_within}
\end{align}
where SC abbreviates Self-ConsU. The sign reflects each component's empirical relationship with accuracy: $\text{SU}_{\text{a}}$ correlates negatively (instability $\to$ failure), so it adds with Self-ConsU; $\text{SU}_{\text{w}}$ correlates positively on reasoning tasks, so it enters subtractively. This assignment is fixed globally---not tuned per task or model. We treat this as an empirical fusion rule rather than an intrinsic part of the decomposition; structural uncertainty captures \emph{how} self-preferences are organized, while Self-ConsU captures \emph{what} answers disagree.

\section{Experiments}
\label{sec:experiments}

\subsection{Experimental Setup}
\label{sec:exp_setup}

We evaluate five LLMs (Claude Sonnet 4.5, GPT-OSS 20B, Qwen 3 32B, Amazon Nova Premier, DeepSeek R1) on eight benchmarks grouped by reasoning structure: \emph{mathematical and logical reasoning} (Math-Synth, MATH-500 \citep{lightman2023lets}, AMC-23 \citep{amc23_2023}, AIME-24/25 \citep{aime_2024,aime25_2024}), \emph{reasoning-adjacent knowledge tasks} (MMLU-Pro \citep{wang2024mmlu}, TruthfulQA \citep{lin2022truthfulqa}), and a \emph{retrieval-dominant comparison regime} (HotpotQA \citep{yang2018hotpotqa}). Math-Synth is a synthetic arithmetic benchmark with 993 problems (Appendix~\ref{app:math_synth}); other dataset details and model accuracies are in Table~\ref{tab:accuracy_all_datasets}. 

We compare against black-box baselines computed from the same $N{=}5$ samples: \textbf{Self-ConsU} (answer entropy) \citep{wang2024self,lyu2025calibrating}, \textbf{SemanticU} (embedding dispersion) \citep{qiu2024semantic,kossen2024semantic}, and \textbf{VerbalizedU} (prompted confidence) \citep{tian2023just,xiong2023can}. See Appendix~\ref{app:baselines} for details.

We use selective prediction metrics as an operational measure of whether the proposed consistency signal identifies unreliable logical reasoning instances: questions where $\hat{p}_{\mathrm{corr}} = \tfrac{1}{N}\sum_i \mathbb{I}[r_i \text{ correct}] < \tau$ with $\tau{=}1.0$. We report Spearman correlation, AUROC, and area under the risk-coverage curve (Sel-AUC) \citep{shrivastava2023llamas}.

\subsection{Results}
\label{sec:results}

We evaluate structural uncertainty as a consistency signal for logical reasoning across five models and eight benchmarks, analyzing where it improves reasoning reliability evaluation, how the components relate to accuracy, and when the signal breaks down. Our key target failure mode is systematic but internally unstable reasoning: candidate solutions may agree at the answer level while remaining inconsistent in how the model ranks them.

\paragraph{Overall Evaluation.}
Table~\ref{tab:main_results} reports reasoning reliability evaluation performance (Sel-AUC; AUROC in parentheses). The central finding is \emph{task-dependent complementarity}: on reasoning-heavy and some knowledge tasks, structural consistency signals add information beyond answer dispersion, while on retrieval-style tasks the structural signal collapses and provides limited benefit. Reasoning tasks admit structurally diverse solution paths, making self-preference consistency informative in logical reasoning regimes. Specifically, the combined estimator (StructU+Self-ConsU) achieves highest or second-highest Sel-AUC on mathematical reasoning (Math-Synth, MATH-500, AMC-23) and knowledge tasks (MMLU-Pro, TruthfulQA), with largest gains where answer-level agreement is insufficient. On HotpotQA, retrieval tasks suppress structural diversity, so the signal collapses---dispersion methods dominate for the two strongest models (Claude: Self-ConsU 0.839 vs.\ combined 0.742; DeepSeek: SemanticU 0.852 vs.\ combined 0.789). This task asymmetry is itself informative: it identifies the regime boundary where reasoning-level consistency evaluation ceases to be useful, making structural uncertainty a regime-sensitive evaluator of reasoning consistency rather than a universal confidence estimator. Among baselines, Self-ConsU is strongest but blind to systematic reasoning errors; VerbalizedU is inconsistent; SemanticU is weakest except where structural signals collapse.

\begin{table*}[t]
\centering
\scriptsize
\setlength{\tabcolsep}{2.0pt}
\renewcommand{\arraystretch}{1.12}
\resizebox{\textwidth}{!}{%
\begin{tabular}{p{1.5cm}p{2.4cm}ccccccccc}
\toprule
\multirow{2}{1.5cm}{Dataset} & \multirow{2}{2.4cm}{Model} &
\multicolumn{3}{c}{\cellcolor{blue!12}\textbf{StructU (Ours)}} &
\multicolumn{3}{c}{\cellcolor{red!12}\textbf{StructU+Self-ConsU (Ours)}} &
\multicolumn{3}{c}{\textbf{Baselines}} \\
\cmidrule(lr){3-5}\cmidrule(lr){6-8}\cmidrule(lr){9-11}
& &
\cellcolor{blue!12}within & \cellcolor{blue!12}across & \cellcolor{blue!12}total &
\cellcolor{red!12}within & \cellcolor{red!12}across & \cellcolor{red!12}total &
Self-ConsU & VerbalizedU & SemanticU \\
\midrule
\multirow{5}{1.4cm}{{Math-Synth}} 
& Claude 4.5 Sonnet & 0.624 (0.984) & 0.644 (0.936) & 0.596 (0.972) & \underline{0.661 (0.992)} & \textbf{0.663 (0.990)} & 0.655 (0.992) & 0.65 (0.989) & 0.544 (0.924) & 0.430 (0.927) \\
& DeepSeek R1 & 0.815 (0.900) & 0.790 (0.837) & 0.773 (0.790) & \textbf{0.823 (0.929)} & \underline{0.820 (0.924)} & 0.814 (0.915) & 0.802 (0.899) & 0.793 (0.701) & 0.664 (0.502) \\
& GPT-OSS 20B & 0.794 (0.756) & 0.762 (0.668) & 0.661 (0.366) & 0.840 (0.955) & \underline{0.849 (0.956)} & \textbf{0.849 (0.958)} & 0.83 (0.958) & 0.792 (0.742) & 0.528 (0.638) \\
& Amazon Nova Premier & 0.496 (0.953) & 0.370 (0.841) & 0.447 (0.853) & \underline{0.511 (0.997)} & 0.498 (0.998) & \textbf{0.512 (0.997)} & 0.382 (0.948) & 0.389 (0.824) & 0.436 (0.807) \\
& Qwen 3 32B & 0.231 (0.794) & 0.190 (0.656) & 0.213 (0.718) & \textbf{0.393 (0.998)} & \underline{0.391 (0.998)} & 0.388 (0.998) & 0.38 (0.995) & 0.279 (0.824) & 0.218 (0.462) \\

\midrule

\multirow{5}{1.4cm}{{MATH-500}} 
& Claude 4.5 Sonnet & 0.931 (0.813) & 0.936 (0.789) & 0.932 (0.801) & 0.947 (0.840) & \textbf{0.950 (0.843)} & \underline{0.948 (0.831)} & 0.942 (0.816) & 0.891 (0.686) & 0.783 (0.720) \\
& DeepSeek R1 & 0.889 (0.652) & 0.890 (0.640) & 0.868 (0.596) & \underline{0.934 (0.776)} & \textbf{0.936 (0.784)} & 0.927 (0.767) & 0.923 (0.759) & 0.870 (0.603) & 0.875 (0.546) \\
& GPT-OSS 20B & 0.869 (0.582) & 0.885 (0.529) & 0.886 (0.611) & 0.897 (0.715) & \textbf{0.910 (0.718)} & \underline{0.906 (0.729)} & 0.871 (0.694) & 0.886 (0.645) & 0.86 (0.46) \\
& Amazon Nova Premier & 0.834 (0.747) & 0.726 (0.609) & 0.821 (0.687) & \textbf{0.887 (0.873)} & 0.880 (0.871) & \underline{0.886 (0.870)} & 0.860 (0.839) & 0.883 (0.715) & 0.81 (0.695) \\
& Qwen 3 32B & 0.859 (0.798) & 0.780 (0.608) & 0.833 (0.754) & \textbf{0.889 (0.882)} & 0.882 (0.853) & \underline{0.888 (0.876)} & 0.871 (0.817) & 0.880 (0.684) & 0.819 (0.717) \\
\midrule

\multirow{5}{1.4cm}{{AMC-23}}
& Claude 4.5 Sonnet & 0.967 (1.000) & 0.966 (0.963) & 0.962 (0.980) & \underline{0.970 (1.000)} & \textbf{0.972 (1.000)} & 0.970 (1.000) & 0.955 (1.00) & 0.900 (0.604) & 0.853 (0.588) \\
& DeepSeek R1 & 0.924 (0.680) & 0.915 (0.463) & 0.923 (0.291) & \textbf{0.985 (1.000)} & \underline{0.985 (1.000)} & 0.985 (1.000) & 0.985 (1.00) & 0.877 (0.412) & 0.880 (0.592) \\
& GPT-OSS 20B & 0.956 (0.770) & 0.931 (0.533) & 0.954 (0.673) & \textbf{0.980 (1.000)} & \underline{0.980 (1.000)} & 0.980 (1.000) & 0.980 (1.00) & 0.980 (0.583) & 0.884 (0.630) \\
& Amazon Nova Premier & 0.591 (0.865) & 0.538 (0.604) & 0.610 (0.919) & 0.712 (1.000) & \underline{0.716 (1.000)} & \textbf{0.718 (1.000)} & 0.584 (1.00) & 0.419 (0.362) & 0.351 (0.410) \\
& Qwen 3 32B & 0.637 (0.643) & 0.566 (0.567) & 0.606 (0.643) & \underline{0.820 (1.000)} & \textbf{0.821 (1.000)} & 0.820 (1.000) & 0.810 (1.00) & 0.637 (0.389) & 0.513 (0.299) \\
\midrule

\multirow{5}{1.4cm}{{AIME-24}} 
& Claude 4.5 Sonnet & 0.497 (0.903) & 0.595 (0.852) & 0.508 (0.875) & \underline{0.662 (0.994)} & \textbf{0.690 (0.989)} & 0.661 (0.994) & 0.567 (0.972) & 0.273 (0.144) & 0.45 (0.64) \\
& DeepSeek R1 & 0.834 (0.752) & 0.820 (0.554) & 0.866 (0.884) & \underline{0.922 (1.000)} & \textbf{0.925 (1.000)} & 0.922 (1.000) & 0.917 (1.00) & 0.799 (0.298) & 0.788 (0.715) \\
& GPT-OSS 20B & 0.604 (0.725) & 0.762 (0.714) & 0.756 (0.813) & 0.893 (1.000) & \textbf{0.907 (1.000)} & 0.896 (1.000) & 0.891 (1.00) & \underline{0.905 (0.319)} & 0.876 (0.681) \\
& Amazon Nova Premier & 0.190 (—) & \underline{0.267 (—)} & 0.170 (—) & 0.181 (—) & 0.210 (—) & 0.189 (—) & 0.174 (--) & 0.125 (--) & \textbf{0.294 (--)} \\
& Qwen 3 32B & 0.235 (—) & 0.243 (—) & 0.257 (—) & 0.401 (—) & \textbf{0.428 (—)} & 0.402 (—) & \underline{0.414 (--)} & 0.355 (--) & 0.213 (1.00) \\
\midrule

\multirow{5}{1.4cm}{{AIME-25}} 
& Claude 4.5 Sonnet & 0.565 (0.975) & 0.602 (0.969) & 0.221 (0.062) & 0.645 (1.00) & \underline{0.646 (1.00)} & \textbf{0.656 (1.00)} & 0.645 (1.00) & 0.489 (0.175) & 0.524 (0.263) \\
& DeepSeek R1 & 0.263 (0.413) & 0.244 (0.259) & 0.539 (0.466) & 0.728 (1.00) & \underline{0.740 (1.00)} & \textbf{0.761 (1.00)} & 0.707 (1.00) & 0.564 (0.296) & 0.616 (0.400) \\
& GPT-OSS 20B & 0.424 (0.449) & 0.392 (0.324) & 0.696 (0.546) & 0.881 (1.00) & \underline{0.884 (1.00)} & \textbf{0.886 (1.00)} & 0.825 (1.00) & 0.776 (0.370) & 0.768 (0.491) \\
& Amazon Nova Premier & 0.203 (--) & 0.149 (--) & 0.059 (--) & 0.198 (--) & \underline{0.218 (--)} & 0.175 (--) & 0.178 (--) & 0.106 (--) & \textbf{0.257 (--)} \\
& Qwen 3 32B & 0.096 (0.862) & 0.179 (0.897) & 0.234 (0.414) & 0.203 (1.00) & \underline{0.282 (1.00)} & \textbf{0.328 (1.00)} & 0.208 (1.00) & 0.192 (0.897) & 0.126 (0.931) \\

\midrule
\multirow{5}{1.4cm}{{MMLU-Pro}}
& Claude 4.5 Sonnet & 0.924 (0.833) & 0.913 (0.785) & 0.916 (0.792) & \textbf{0.944 (0.912)} & 0.936 (0.897) & 0.943 (0.908) & 0.900 (0.884) & \underline{0.944 (0.885)} & 0.890 (0.602) \\
& DeepSeek R1 & 0.845 (0.573) & 0.849 (0.500) & 0.855 (0.596) & \underline{0.925 (0.889)} & 0.917 (0.882) & 0.924 (0.895) & 0.882 (0.882) & \textbf{0.927 (0.796)} & 0.870 (0.577) \\
& GPT-OSS 20B & 0.765 (0.503) & 0.754 (0.547) & 0.749 (0.475) & \textbf{0.889 (0.948)} & 0.880 (0.941) & \underline{0.886 (0.945)} & 0.830 (0.935) & 0.785 (0.631) & 0.774 (0.573) \\
& Amazon Nova Premier & 0.671 (0.633) & 0.696 (0.507) & 0.679 (0.624) & 0.805 (0.936) & \textbf{0.827 (0.948)} & 0.800 (0.936) & 0.801 (0.945) & \underline{0.820 (0.775)} & 0.811 (0.713) \\
& Qwen 3 32B & 0.713 (0.669) & 0.657 (0.567) & 0.685 (0.624) & \textbf{0.818 (0.971)} & 0.805 (0.964) & \underline{0.817 (0.971)} & 0.787 (0.966) & 0.728 (0.724) & 0.742 (0.686) \\

\midrule
\multirow{5}{1.4cm}{{HotpotQA}} 
& Claude 4.5 Sonnet & 0.686 (0.585) & 0.731 (0.600) & 0.664 (0.564) & 0.698 (0.617) & 0.742 (0.647) & 0.681 (0.604) & \underline{0.839 (0.656)} & \textbf{0.847 (0.700)} & 0.768 (0.576) \\
& DeepSeek R1 & 0.732 (0.580) & 0.747 (0.514) & 0.744 (0.614) & 0.767 (0.683) & 0.789 (0.666) & 0.771 (0.693) & \underline{0.835 (0.658)} & 0.829 (0.708) & \textbf{0.852 (0.696)} \\
& GPT-OSS 20B & 0.724 (0.537) & 0.717 (0.524) & 0.736 (0.568) & \underline{0.815 (0.728)} & 0.798 (0.724) & \textbf{0.815 (0.734)} & 0.813 (0.721) & 0.772 (0.592) & 0.80 (0.707) \\
& Amazon Nova Premier & 0.806 (0.582) & 0.805 (0.589) & 0.787 (0.541) & 0.850 (0.752) & 0.840 (0.748) & \underline{0.855 (0.750)} & \textbf{0.864 (0.740)} & 0.812 (0.647) & 0.830 (0.618) \\
& Qwen 3 32B & 0.699 (0.648) & 0.679 (0.528) & 0.691 (0.618) & 0.766 (0.757) & 0.772 (0.713) & \underline{0.778 (0.765)} & \textbf{0.787 (0.729)} & 0.702 (0.630) & 0.707 (0.612) \\
\midrule

\multirow{5}{1.4cm}{{TruthfulQA}}  
& Claude 4.5 Sonnet & 0.998 (0.956) & 0.997 (0.912) & 0.998 (0.953) & \underline{0.998 (0.970)} & 0.998 (0.966) & \textbf{0.998 (0.971)} & 0.994 (0.926) & 0.997 (0.949) & 0.979 (0.503) \\
& DeepSeek R1 & 0.932 (0.564) & 0.941 (0.662) & 0.874 (0.535) & \underline{0.954 (0.884)} & \textbf{0.957 (0.888)} & 0.910 (0.813) & 0.940 (0.844) & 0.949 (0.721) & 0.907 (0.530) \\
& GPT-OSS 20B & 0.913 (0.717) & 0.893 (0.611) & 0.909 (0.727) & \textbf{0.943 (0.955)} & \underline{0.942 (0.955)} & 0.941 (0.952) & 0.916 (0.936) & 0.864 (0.608) & 0.856 (0.525) \\
& Amazon Nova Premier & 0.956 (0.806) & 0.930 (0.689) & 0.952 (0.779) & \textbf{0.963 (0.914)} & 0.951 (0.893) & \underline{0.962 (0.914)} & 0.946 (0.862) & 0.953 (0.828) & 0.907 (0.596) \\
& Qwen 3 32B & 0.783 (0.510) & 0.787 (0.534) & 0.765 (0.499) & \textbf{0.869 (0.956)} & \underline{0.867 (0.953)} & 0.855 (0.949) & 0.856 (0.936) & 0.822 (0.694) & 0.812 (0.636) \\
\bottomrule
\end{tabular}%
}
\caption{\textbf{Selective prediction performance (Sel-AUC; AUROC in parentheses).} 
Results using Bradley--Terry with PageRank aggregation. 
StructU reports structural uncertainty (within, across components). 
StructU+Self-ConsU reports the combined estimators. 
Baselines: Self-ConsU (answer entropy), VerbalizedU (prompted confidence), SemanticU (embedding dispersion). 
Bold = best; underline = second-best per row.}
\label{tab:main_results}
\end{table*}

\begin{figure}[t]
    \centering
    \includegraphics[width=\linewidth]{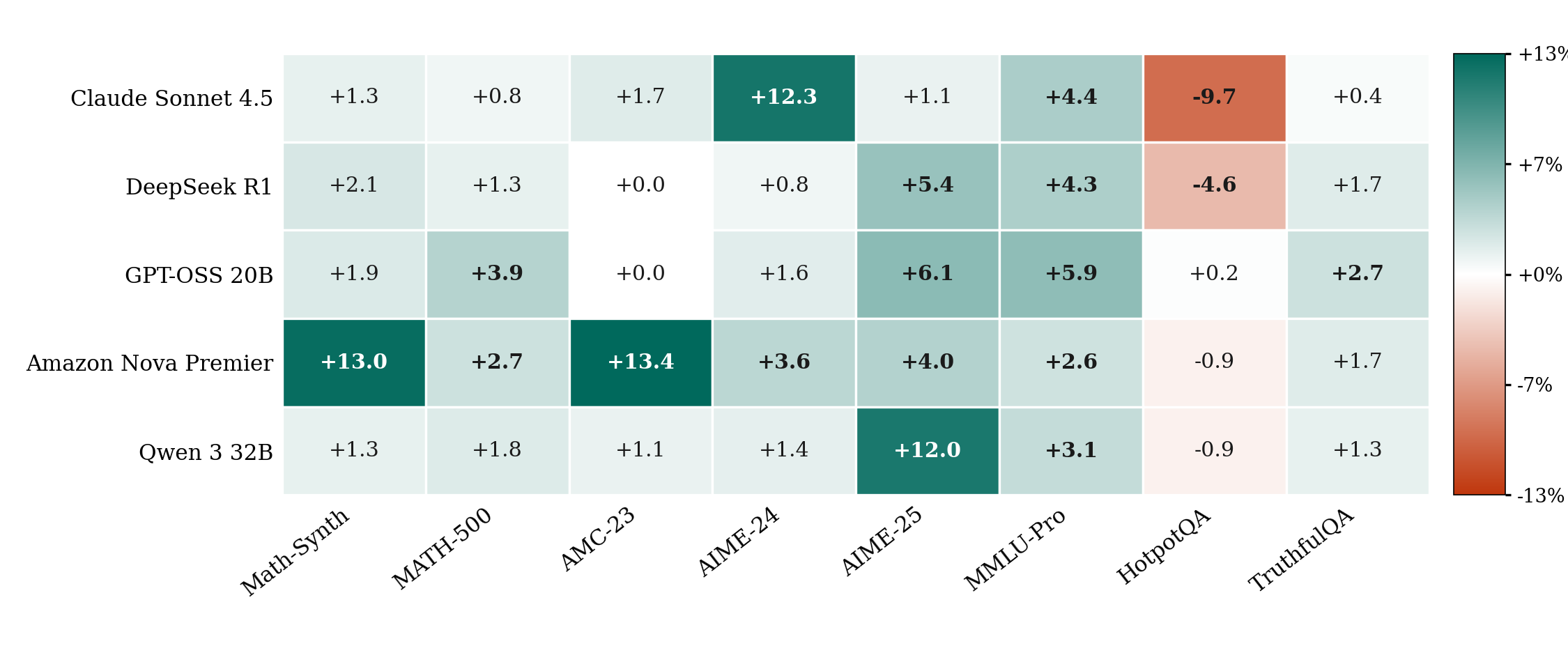}
    \caption{\textbf{Sel-AUC lift of \textsc{StructU+Self-ConsU} over \textsc{Self-ConsU}.}
$\Delta\text{Sel-AUC}$ for five models $\times$ eight benchmarks.
Teal = hybrid wins; coral = Self-ConsU dominates.}
\label{fig:heatmap_selauc}
\end{figure}

\paragraph{Where Structural Consistency Signals Help.}
Figure~\ref{fig:heatmap_selauc} shows $\Delta\text{Sel-AUC} = \text{Sel-AUC}(\textsc{StructU+Self-ConsU}) - \text{Sel-AUC}(\text{Self-ConsU})$.
Gains are consistent across logical reasoning and knowledge benchmarks, and largest where answer-level agreement alone is insufficient to identify unreliable reasoning: weaker models on hard contest problems gain the most (Qwen on AIME-25: $+12.0\%$, Amazon Nova Premier on AMC-23: $+13.4\%$), as structural consistency rankings surface signal not captured by Self-ConsU alone.
HotpotQA is the exception---stronger models show negative lift (Claude Sonnet 4.5: $-9.7\%$, DeepSeek R1: $-4.6\%$), consistent with the structural collapse on retrieval tasks.

\begin{figure*}[!htbp]
    \centering
    \includegraphics[width=0.85\textwidth]{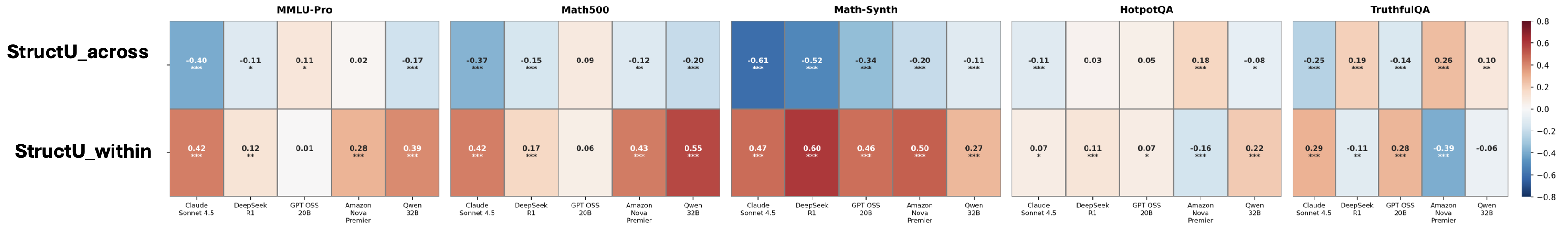}
    \caption{\textbf{Spearman correlation ($\rho$) between the two uncertainty components and accuracy.}
    Each cell reports Spearman rank correlation between per-question uncertainty and per-question accuracy (fraction correct among $N{=}5$ samples).
    Blue indicates negative correlation; red indicates positive correlation.
    Stars denote statistical significance ($^{*}p < 0.05$, $^{**}p < 0.01$, $^{***}p < 0.001$).}
    \label{fig:rho_accuracy_heatmap}
\end{figure*}

\paragraph{Correlation Between Consistency Signal and Reasoning Accuracy.}
Figure~\ref{fig:rho_accuracy_heatmap} shows across-trial and within-trial components exhibit opposite correlations with accuracy---most pronounced on mathematical reasoning. On Math-Synth and MATH-500, across-trial instability is negatively correlated with correctness while within-trial ambiguity is positively correlated (e.g., Claude on MATH-500: $\rho_{\text{across}}=-0.37$ vs. $\rho_{\text{within}}=0.42$). This asymmetry has a natural interpretation for logical reasoning: ranking instability signals unreliable reasoning, while distributed preference among candidates is consistent with settings where multiple plausible solution paths remain competitive. The pattern weakens on MMLU-Pro and collapses on HotpotQA (near-zero correlations), confirming that the consistency signal is regime-sensitive---informative for logical reasoning but uninformative where reasoning-level structural diversity is absent.

\begin{figure*}[!t]
\centering
\includegraphics[width=0.8\textwidth]{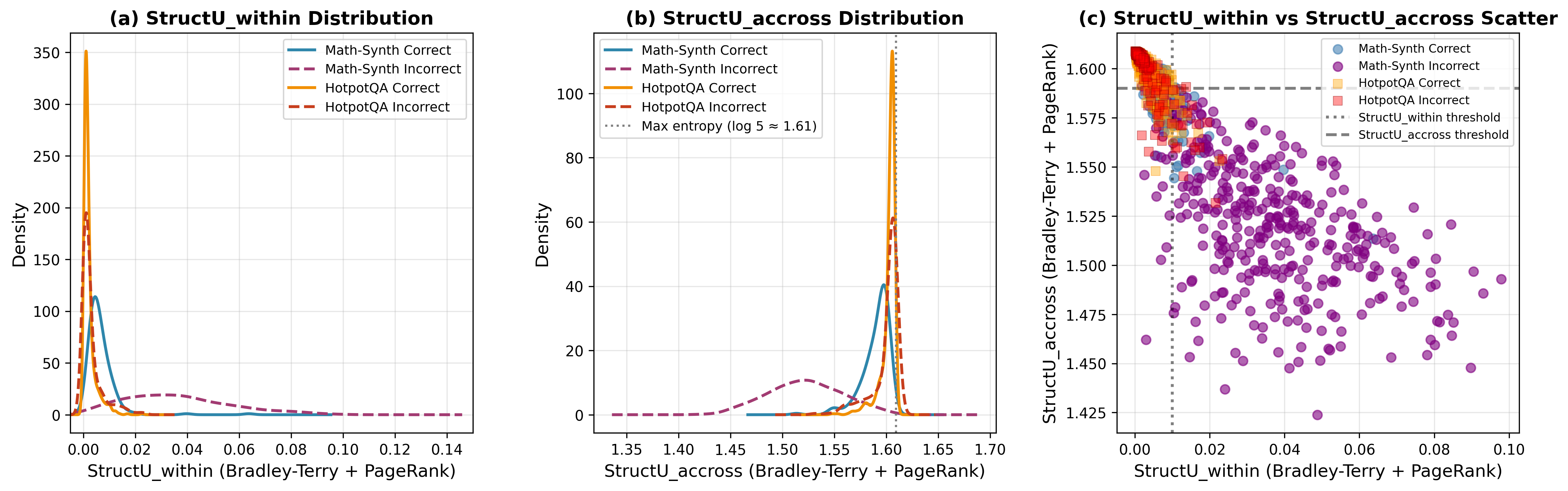}
\caption{\textbf{Regime analysis: reasoning consistency vs.\ retrieval collapse.}
Claude~4.5~Sonnet on Math-Synth (reasoning) and HotpotQA (retrieval), conditioned on correctness (BT + PageRank).
\textbf{(a)}~\textsc{StructU}\textsubscript{across}: Math-Synth shows correctness separation; HotpotQA concentrates near zero for both.
\textbf{(b)}~\textsc{StructU}\textsubscript{within}: HotpotQA clusters at maximum entropy ($\log 5 \approx 1.61$, dotted line).
\textbf{(c)}~Joint space: reasoning tasks separate along the across-trial axis; retrieval tasks collapse into a degenerate cluster, diagnosing the regime boundary where logical reasoning structure is absent.}
\label{fig:structural_collapse}
\end{figure*}

\paragraph{Regime Analysis: Reasoning vs.\ Retrieval.}
Table~\ref{tab:main_results} reveals a clear regime boundary: StructU+Self-ConsU consistently outperforms baselines on logical reasoning, whereas on HotpotQA, dispersion methods dominate for the strongest models (e.g., Claude: Self-ConsU 0.839 vs. Hybrid 0.742; DeepSeek: SemanticU 0.852 vs. Hybrid 0.796). Understanding \emph{when} consistency signals fail is as important as knowing when they succeed, as this clarifies where logical reasoning structure is present versus absent. Figure~\ref{fig:structural_collapse} compares Claude 4.5 Sonnet on Math-Synth and HotpotQA under identical experimental conditions ($N{=}5$, $M{=}5$, similar prompt templates).

\noindent\textbf{Across-trial separation on reasoning, collapse on retrieval.}
On Math-Synth, across-trial uncertainty distributions are well-separated between correct and incorrect questions (Figure~\ref{fig:structural_collapse}a): incorrect questions exhibit a long right tail beyond 0.10, while correct questions concentrate near zero. On HotpotQA, both distributions are compressed near zero---structural uncertainty produces no across-trial signal regardless of correctness. In the joint space (Figure~\ref{fig:structural_collapse}c), Math-Synth shows correctness separation along the across-trial axis, while HotpotQA collapses into a degenerate cluster with no separation.

\noindent\textbf{Near-uniform preference distributions on factual retrieval.}
The within-trial distributions (Figure~\ref{fig:structural_collapse}b) reveal the mechanism: HotpotQA responses cluster at maximum entropy ($\log 5 \approx 1.61$, dotted line) for both correct and incorrect questions, indicating near-uniform PageRank distributions. The model assigns approximately equal preference to all five responses, producing pairwise confidences near 50\% and identical rankings across spanning tree samples.

\noindent\textbf{When does self-preference carry signal?}
Logical and mathematical reasoning elicits \emph{structurally diverse} solution paths---step-by-step computation, bracket decomposition, estimation-then-verification---enabling meaningful self-preference discrimination and making the consistency signal informative. In our setting, factual retrieval often elicits more \emph{structurally homogeneous} reasoning chains across prompt templates, so self-preference reflects only stylistic variation and the structural signal vanishes. The ``HotpotQA signature'' in Figure~\ref{fig:structural_collapse}c serves as a practical diagnostic: it identifies the regime boundary where logical reasoning structure is absent and dispersion-based methods should be preferred. This collapse is not merely a failure case; it is a substantively useful boundary result that distinguishes tasks genuinely supporting reasoning-consistency analysis from those dominated by retrieval-induced homogeneity.

\begin{table}[t]
\centering
\small
\setlength{\tabcolsep}{3.5pt}
\renewcommand{\arraystretch}{1.12}
\resizebox{\columnwidth}{!}{%
\begin{tabular}{lcccccc}
\toprule
\multirow{2}{*}{Model} &
\multicolumn{3}{c}{\textbf{\textsc{StructU}$_{\text{within}}$ (Within-trial) AUROC}} &
\multicolumn{3}{c}{\textbf{\textsc{StructU}$_{\text{across}}$ (Across-trial) AUROC}} \\
\cmidrule(lr){2-4}\cmidrule(lr){5-7}
& Real & Random & $\downarrow$ Drop &
  Real & Random & $\downarrow$ Drop \\
\midrule
Claude 4.5 Sonnet    & 0.984 & 0.488 $\pm$ 0.016 & 0.496 & 0.923 & 0.510 $\pm$ 0.019 & 0.413 \\
DeepSeek R1          & 0.896 & 0.743 $\pm$ 0.005 & 0.153 & 0.840 & 0.747 $\pm$ 0.001 & 0.093 \\
GPT-OSS 20B          & 0.756 & 0.557 $\pm$ 0.048 & 0.199 & 0.668 & 0.588 $\pm$ 0.054 & 0.080 \\
Amazon Nova Premier  & 0.943 & 0.530 $\pm$ 0.005 & 0.413 & 0.866 & 0.482 $\pm$ 0.014 & 0.384 \\
Qwen 3 32B           & 0.850 & 0.511 $\pm$ 0.021 & 0.339 & 0.732 & 0.512 $\pm$ 0.020 & 0.220 \\
\midrule
\textbf{Mean}        & \textbf{0.886} & \textbf{0.566 $\pm$ 0.022} & \textbf{0.320} &
                       \textbf{0.806} & \textbf{0.568 $\pm$ 0.022} & \textbf{0.238} \\
\bottomrule
\end{tabular}%
}
\caption{\textbf{Real vs.\ randomized preferences on Math-Synth (BT+PageRank).}
Randomization tests whether gains depend on elicited self-preference content rather than fixed aggregation structure: winner direction and confidence scores are randomized while all other pipeline components
are held fixed. Random AUROC is mean~$\pm$~std over three runs;
$\downarrow$ indicates drop from real to random.}
\label{tab:random_ablation}
\end{table}

\begin{figure}[!h]
  \centering
  \includegraphics[width=0.7\linewidth]{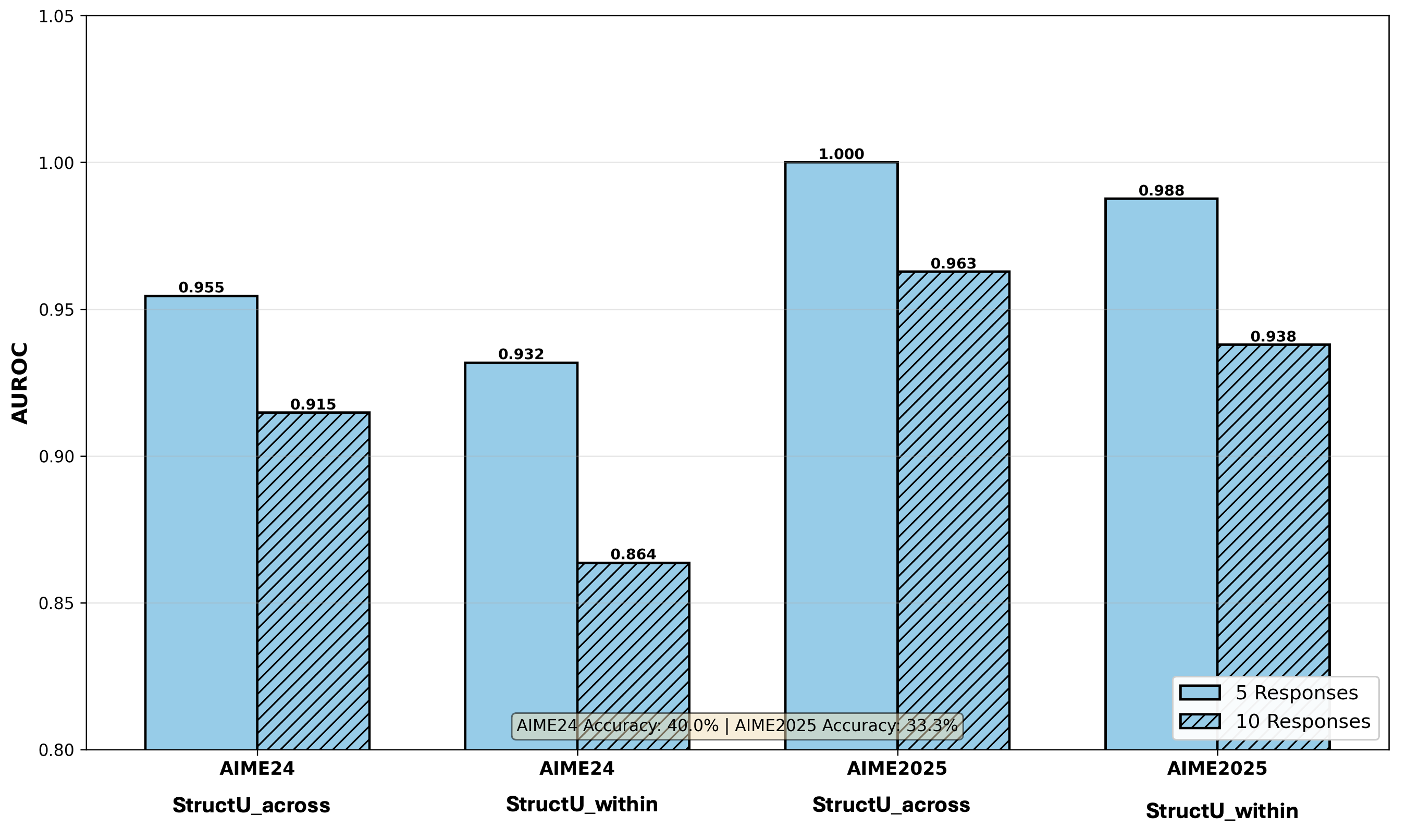}
  \caption{Ablation study on AIME benchmarks: effect of number of sampled responses showing performance degrades with $N{=}10$.}
  \label{fig:ablation_responses}
\end{figure}

\begin{figure}[!h]
  \centering
  \includegraphics[width=0.7\linewidth]{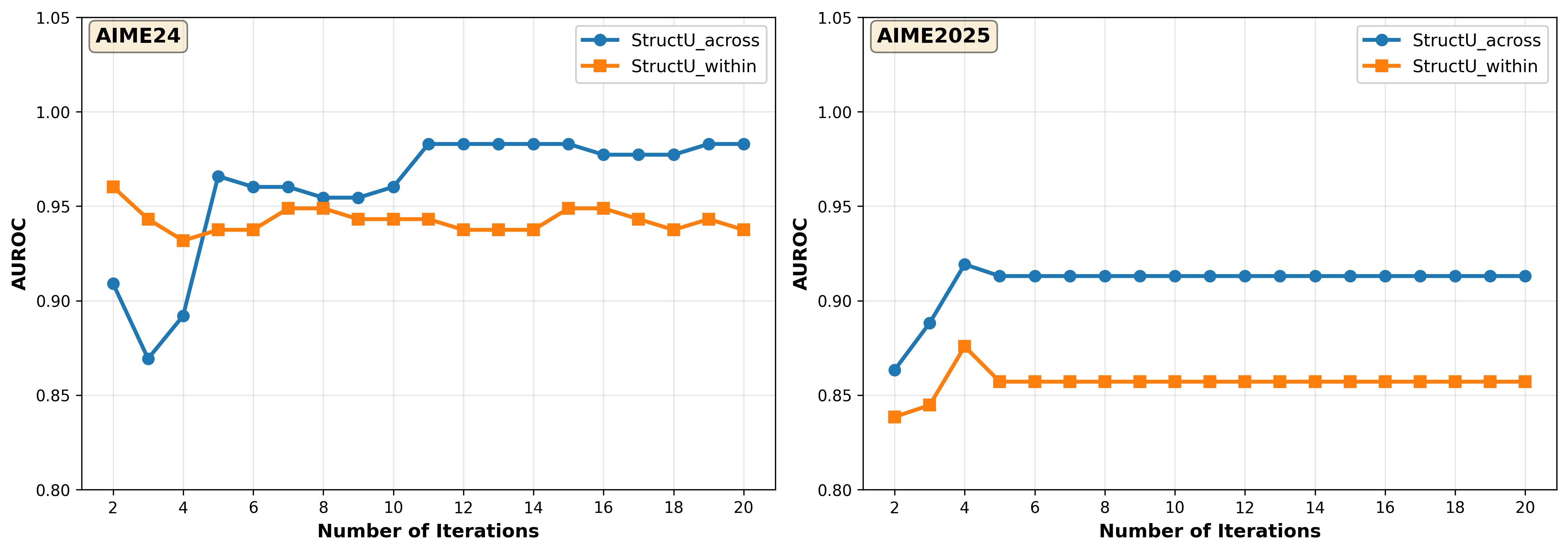}
  \caption{Ablation study on AIME benchmarks: effect of number of trials showing performance plateaus at $M{\approx}5$.}
  \label{fig:ablation_iterations}
\end{figure}

\paragraph{Ablation Studies.}
To test whether structural uncertainty reflects elicited preference signal rather than pipeline artifacts, we replace real self-preference judgments with random comparisons on Math-Synth: winner direction is randomized uniformly and confidence scores are sampled uniformly from $[51, 99]$, while all other pipeline components (spanning tree topology, BT fitting with $C{=}1$, PageRank aggregation, entropy computation) are held fixed. Table~\ref{tab:random_ablation} shows AUROC drops substantially for both uncertainty components (mean drop: 0.320 for within-trial, 0.238 for across-trial), with three models collapsing to near-chance level (Claude: 0.984 → 0.488; Nova: 0.943 → 0.530; Qwen: 0.850 → 0.511). These ablations suggest that discriminative performance depends materially on the elicited self-preference judgments rather than on the aggregation structure alone.
Performance plateaus at $M{\approx}5$ trials and remains stable through $M{=}20$ (Figure~\ref{fig:ablation_iterations}). The $C{=}1$ regularization choice is validated as performance degrades for $C < 1$ but remains stable for $C \geq 1$ across all models. PageRank smoothing provides mild regularization benefit ($+0.015$ Sel-AUC average). Increasing response count from $N{=}5$ to $N{=}10$ degrades performance, indicating high-temperature samples introduce noise (Figure~\ref{fig:ablation_responses}).

\section{Conclusion}
We introduced structural uncertainty, a consistency-aware evaluation framework for logical reasoning that measures the stability of self-preference-induced rankings over sampled LLM reasoning candidates. By eliciting pairwise self-preferences and aggregating them via Bradley--Terry with PageRank, we decompose the signal into across-trial ranking instability and within-trial candidate ambiguity without requiring model internals.

Across five LLMs and eight benchmarks, the central finding is that structural uncertainty provides a consistency-aware lens on logical reasoning: combining structural and dispersion-based signals improves identification of unreliable reasoning instances on several reasoning and knowledge tasks, with largest gains where systematic errors produce consistent but incorrect answers---a failure mode invisible to answer dispersion alone. The two components relate differently to accuracy---across-trial instability signals unreliable reasoning, while within-trial ambiguity correlates positively on mathematical reasoning, consistent with settings where multiple plausible solution paths remain competitive. Conversely, on factual retrieval (HotpotQA), the structural signal collapses where reasoning-level structural diversity is absent; this collapse itself clarifies the regime boundary where logical reasoning structure is not present.

Structural uncertainty is best understood not as a universal confidence estimator, but as a regime-sensitive evaluator of logical reasoning consistency---reframing the question from \emph{how much do responses differ} to \emph{how consistently does the model rank competing reasoning solutions}. These two complementary views of the same response set help practitioners assess reasoning reliability and identify when consistency-based evaluation is informative versus when answer dispersion should be preferred. More broadly, our results suggest that benchmarking logical reasoning should account not only for answer agreement, but also for the structural stability of model-internal preferences over competing solution paths.

\section*{Limitations}
Our approach requires $N$ generations and $M(N{-}1)$ pairwise comparisons per question ($N{=}5$, $M{=}5$), increasing inference cost. Since models judge their own outputs, self-preferences can inherit model-specific biases and should be interpreted as behavioral signals rather than guaranteed correctness measures. The decomposition is empirical: the two components are observable signals under a fixed protocol, not universally identifiable uncertainty sources---candidate diversity, prompt design, and task structure all affect the signal. The method is most informative when responses differ in reasoning quality; when variation is stylistic, the preference graph collapses (as on HotpotQA). Accordingly, our method should be interpreted as an evaluator of reasoning consistency under a fixed elicitation protocol, rather than as a complete measure of logical validity. Our evaluation focuses on short-answer tasks; extending to long-form or open-ended generation remains future work.

\FloatBarrier

\bibliographystyle{plainnat}
\bibliography{structu}

@article{abbasi2024believe,
  title={To believe or not to believe your llm: Iterative prompting for estimating epistemic uncertainty},
  author={Abbasi Yadkori, Yasin and Kuzborskij, Ilja and Gy{\"o}rgy, Andr{\'a}s and Szepesvari, Csaba},
  journal={Advances in Neural Information Processing Systems},
  volume={37},
  pages={58077--58117},
  year={2024}
}

@inproceedings{lyu2025calibrating,
  title={Calibrating Large Language Models with Sample Consistency},
  author={Lyu, Qing and Shridhar, Kumar and Malaviya, Chaitanya and Zhang, Li and Elazar, Yanai and Tandon, Niket and Apidianaki, Marianna and Sachan, Mrinmaya and Callison-Burch, Chris},
  booktitle={Proceedings of the AAAI Conference on Artificial Intelligence},
  volume={39},
  pages={19260--19268},
  year={2025}
}

@article{lin2023generating,
  title={Generating with confidence: Uncertainty quantification for black-box large language models},
  author={Lin, Zhen and Trivedi, Shubhendu and Sun, Jimeng},
  journal={arXiv preprint arXiv:2305.19187},
  year={2023}
}

@article{farquhar2024detecting,
  title={Detecting hallucinations in large language models using semantic entropy},
  author={Farquhar, Sebastian and Kossen, Jannik and Kuhn, Lorenz and Gal, Yarin},
  journal={Nature},
  volume={630},
  number={8017},
  pages={625--630},
  year={2024},
  publisher={Nature Publishing Group UK London}
}

@article{kossen2024semantic,
  title={Semantic entropy probes: Robust and cheap hallucination detection in llms},
  author={Kossen, Jannik and Han, Jiatong and Razzak, Muhammed and Schut, Lisa and Malik, Shreshth and Gal, Yarin},
  journal={arXiv preprint arXiv:2406.15927},
  year={2024}
}

@article{qiu2024semantic,
  title={Semantic density: Uncertainty quantification for large language models through confidence measurement in semantic space},
  author={Qiu, Xin and Miikkulainen, Risto},
  journal={Advances in neural information processing systems},
  volume={37},
  pages={134507--134533},
  year={2024}
}

@article{nikitin2024kernel,
  title={Kernel language entropy: Fine-grained uncertainty quantification for llms from semantic similarities},
  author={Nikitin, Alexander and Kossen, Jannik and Gal, Yarin and Marttinen, Pekka},
  journal={Advances in Neural Information Processing Systems},
  volume={37},
  pages={8901--8929},
  year={2024}
}

@article{shrivastava2025language,
  title={Language Models Prefer What They Know: Relative Confidence Estimation via Confidence Preferences},
  author={Shrivastava, Vaishnavi and Kumar, Ananya and Liang, Percy},
  journal={arXiv preprint arXiv:2502.01126},
  year={2025}
}

@inproceedings{yuan2025kg,
  title={KG-UQ: Knowledge Graph-Based Uncertainty Quantification for Long Text in Large Language Models},
  author={Yuan, Yingqing and Tao, Linwei and Lu, Haohui and Khushi, Matloob and Razzak, Imran and Dras, Mark and Yang, Jian and Naseem, Usman},
  booktitle={Companion Proceedings of the ACM on Web Conference 2025},
  pages={2071--2077},
  year={2025}
}

@article{chen2025uncertainty,
  title={Uncertainty Quantification of Large Language Models through Multi-Dimensional Responses},
  author={Chen, Tiejin and Liu, Xiaoou and Da, Longchao and Chen, Jia and Papalexakis, Vagelis and Wei, Hua},
  journal={arXiv preprint arXiv:2502.16820},
  year={2025}
}

@article{ma2025estimating,
  title={Estimating LLM Uncertainty with Evidence},
  author={Ma, Huan and Chen, Jingdong and Zhou, Joey Tianyi and Wang, Guangyu and Zhang, Changqing},
  journal={arXiv preprint arXiv:2502.00290},
  year={2025}
}

@inproceedings{zhang2025cot,
  title={{CoT-UQ}: Improving Response-Wise Uncertainty Quantification in {LLMs} with Chain-of-Thought},
  author={Zhang, Boxuan and Zhang, Ruqi},
  booktitle={Findings of the Association for Computational Linguistics: ACL 2025},
  year={2025},
  publisher={Association for Computational Linguistics}
}

@InProceedings{lee2024improving,
  title = 	 {Improving Instruction Following in Language Models through Proxy-Based Uncertainty Estimation},
  author =       {Lee, Joonho and Woo, Jae Oh and Seok, Juree and Hassanzadeh, Parisa and Jang, Wooseok and Son, Juyoun and Didari, Sima and Gutow, Baruch and Hao, Heng and Moon, Hankyu and Hu, Wenjun and Kwon, Yeong-Dae and Lee, Taehee and Min, Seungjai},
  booktitle = 	 {Proceedings of the 41st International Conference on Machine Learning},
  pages = 	 {27009--27036},
  year = 	 {2024},
  volume = 	 {235},
  series = 	 {Proceedings of Machine Learning Research},
  month = 	 {21--27 Jul},
  publisher =    {PMLR}
}

@inproceedings{kapoor2024calibration,
  title={Calibration-tuning: Teaching large language models to know what they don’t know},
  author={Kapoor, Sanyam and Gruver, Nate and Roberts, Manley and Pal, Arka and Dooley, Samuel and Goldblum, Micah and Wilson, Andrew},
  booktitle={Proceedings of the 1st Workshop on Uncertainty-Aware NLP (UncertaiNLP 2024)},
  pages={1--14},
  year={2024}
}

@article{shrivastava2023llamas,
  title={Llamas Know What GPTs Don't Show: Surrogate Models for Confidence Estimation},
  author={Shrivastava, Vaishnavi and Liang, Percy and Kumar, Ananya},
  journal={arXiv preprint arXiv:2311.08877},
  year={2023}
}

@inproceedings{tian2023just,
  title={Just Ask for Calibration: Strategies for Eliciting Calibrated Confidence Scores from Language Models Fine-Tuned with Human Feedback},
  author={Tian, Katherine and Mitchell, Eric and Zhou, Allan and Sharma, Archit and Rafailov, Rafael and Yao, Huaxiu and Finn, Chelsea and Manning, Christopher D.},
  booktitle={Proceedings of the 2023 Conference on Empirical Methods in Natural Language Processing},
  pages={5433--5442},
  year={2023},
  publisher={Association for Computational Linguistics}
}

@inproceedings{huang2024uncertainty,
  title={Uncertainty in Language Models: Assessment through Rank-Calibration},
  author={Huang, Xinmeng and Li, Shuo and Yu, Mengxin and Sesia, Matteo and Hassani, Hamed and Lee, Insup and Bastani, Osbert and Dobriban, Edgar},
  booktitle={Proceedings of the 2024 Conference on Empirical Methods in Natural Language Processing},
  pages={2851--2873},
  year={2024},
  publisher={Association for Computational Linguistics}
}

@article{ye2024benchmarking,
  title={Benchmarking llms via uncertainty quantification},
  author={Ye, Fanghua and Yang, Mingming and Pang, Jianhui and Wang, Longyue and Wong, Derek and Yilmaz, Emine and Shi, Shuming and Tu, Zhaopeng},
  journal={Advances in Neural Information Processing Systems},
  volume={37},
  pages={15356--15385},
  year={2024}
}

@article{huang2023look,
  title={Look before you leap: An exploratory study of uncertainty measurement for large language models},
  author={Huang, Yuheng and Song, Jiayang and Wang, Zhijie and Zhao, Shengming and Chen, Huaming and Juefei-Xu, Felix and Ma, Lei},
  journal={arXiv preprint arXiv:2307.10236},
  year={2023}
}

@article{wang2024self,
  title={Self-consistency boosts calibration for math reasoning},
  author={Wang, Ante and Song, Linfeng and Tian, Ye and Peng, Baolin and Jin, Lifeng and Mi, Haitao and Su, Jinsong and Yu, Dong},
  journal={arXiv preprint arXiv:2403.09849},
  year={2024}
}

@inproceedings{woo2023active,
title={Active Learning in Bayesian Neural Networks with Balanced Entropy Learning Principle},
author={Jae Oh Woo},
booktitle={The Eleventh International Conference on Learning Representations },
year={2023},
url={https://openreview.net/forum?id=ZTMuZ68B1g}
}

@inproceedings{yang2025maqa,
  title={Maqa: Evaluating uncertainty quantification in llms regarding data uncertainty},
  author={Yang, Yongjin and Yoo, Haneul and Lee, Hwaran},
  booktitle={Findings of the Association for Computational Linguistics: NAACL 2025},
  pages={5846--5863},
  year={2025}
}

@article{vashurin2025benchmarking,
  title={Benchmarking uncertainty quantification methods for large language models with lm-polygraph},
  author={Vashurin, Roman and Fadeeva, Ekaterina and Vazhentsev, Artem and Rvanova, Lyudmila and Vasilev, Daniil and Tsvigun, Akim and Petrakov, Sergey and Xing, Rui and Sadallah, Abdelrahman and Grishchenkov, Kirill and others},
  journal={Transactions of the Association for Computational Linguistics},
  volume={13},
  pages={220--248},
  year={2025},
  publisher={MIT Press}
}

@inproceedings{vashurin2025cocoa,
  title={{CoCoA}: A Minimum {B}ayes Risk Framework Bridging Confidence and Consistency for Uncertainty Quantification in Large Language Models},
  author={Vashurin, Alexey and Vikhreva, Maria and Kocmi, Tom and Malinin, Andrey},
  booktitle={Advances in Neural Information Processing Systems},
  volume={38},
  year={2025}
}

@article{BradleyTerry1952,
  author  = {Bradley, Ralph Allan and Terry, Milton E.},
  title   = {Rank Analysis of Incomplete Block Designs: {I}. The Method of Paired Comparisons},
  journal = {Biometrika},
  volume  = {39},
  number  = {3-4},
  pages   = {324--345},
  year    = {1952},
  month   = {12},
  doi     = {10.1093/biomet/39.3-4.324}
}

@article{Hunter2004MMBradleyTerry,
  author  = {Hunter, David R.},
  title   = {{MM} Algorithms for Generalized {Bradley--Terry} Models},
  journal = {The Annals of Statistics},
  volume  = {32},
  number  = {1},
  pages   = {384--406},
  year    = {2004},
  month   = {2},
  doi     = {10.1214/aos/1079120141}
}

@inproceedings{BrinPage1998Anatomy,
  author    = {Brin, Sergey and Page, Lawrence},
  title     = {The Anatomy of a Large-Scale Hypertextual {W}eb Search Engine},
  booktitle = {Computer Networks and {ISDN} Systems},
  volume    = {30},
  pages     = {107--117},
  year      = {1998},
  doi       = {10.1016/S0169-7552(98)00110-X}
}

@book{LangvilleMeyer2006PageRank,
  author    = {Langville, Amy N. and Meyer, Carl D.},
  title     = {Google's {P}age{R}ank and Beyond: The Science of Search Engine Rankings},
  publisher = {Princeton University Press},
  year      = {2006},
  isbn      = {0691122024}
}

@inproceedings{KendallGal2017Uncertainties,
  author    = {Kendall, Alex and Gal, Yarin},
  title     = {What Uncertainties Do We Need in {B}ayesian Deep Learning for Computer Vision?},
  booktitle = {Advances in Neural Information Processing Systems ({NeurIPS})},
  year      = {2017},
  url       = {https://arxiv.org/abs/1703.04977}
}

@inproceedings{xiong2023can,
  title={Can {LLMs} Express Their Uncertainty? An Empirical Evaluation of Confidence Elicitation in {LLMs}},
  author={Xiong, Miao and Hu, Zhiyuan and Lu, Xinyang and Li, Yifei and Fu, Jie and He, Junxian and Hooi, Bryan},
  booktitle={Proceedings of the Twelfth International Conference on Learning Representations},
  year={2024}
}

@inproceedings{woo2022analytic,
  title={Analytic mutual information in bayesian neural networks},
  author={Woo, Jae Oh},
  booktitle={2022 IEEE International Symposium on Information Theory (ISIT)},
  pages={300--305},
  year={2022},
  organization={IEEE}
}

@article{zhou2024relying,
  title={Relying on the unreliable: The impact of language models' reluctance to express uncertainty},
  author={Zhou, Kaitlyn and Hwang, Jena D and Ren, Xiang and Sap, Maarten},
  journal={arXiv preprint arXiv:2401.06730},
  year={2024}
}

@inproceedings{Wilson1996RandomSpanningTrees,
  author    = {Wilson, David Bruce},
  title     = {Generating Random Spanning Trees More Quickly than the Cover Time},
  booktitle = {Proceedings of the Twenty-Eighth Annual ACM Symposium on Theory of Computing},
  series    = {STOC '96},
  year      = {1996},
  pages     = {296--303},
  publisher = {Association for Computing Machinery},
  address   = {New York, NY, USA},
  doi       = {10.1145/237814.237880}
}

@article{herbrich2006trueskill,
  title={TrueSkill™: a Bayesian skill rating system},
  author={Herbrich, Ralf and Minka, Tom and Graepel, Thore},
  journal={Advances in neural information processing systems},
  volume={19},
  year={2006}
}

@article{Ford1957,
  author  = {Ford, L. R.},
  title   = {Solution of a Ranking Problem from Binary Comparisons},
  journal = {The American Mathematical Monthly},
  year    = {1957},
  volume  = {64},
  number  = {8},
  pages   = {28--33}
}

@misc{aime_2024,
  title = {{AIME 2024}: American Invitational Mathematics Examination Dataset},
  author = {HuggingFaceH4},
  year = {2024},
  howpublished = {HuggingFace Dataset},
  url = {https://huggingface.co/datasets/HuggingFaceH4/aime_2024}
}

@misc{aime25_2024,
  title = {{AIME25}: A Benchmark for Mathematical Reasoning},
  author = {TIGER-Lab},
  year = {2024},
  howpublished = {HuggingFace Dataset},
  url = {https://huggingface.co/datasets/TIGER-Lab/AIME25}
}

@misc{amc23_2023,
  title = {{AMC23}: American Mathematics Competitions Dataset},
  author = {He, Zhiwei},
  year = {2023},
  howpublished = {HuggingFace Dataset},
  url = {https://huggingface.co/datasets/zwhe99/amc23}
}

@inproceedings{lightman2023lets,
  title={Let's Verify Step by Step},
  author={Lightman, Hunter and Kosaraju, Vineet and Burda, Yura and Edwards, Harri and Baker, Bowen and Lee, Teddy and Leike, Jan and Schulman, John and Sutskever, Ilya and Cobbe, Karl},
  booktitle={Proceedings of the Twelfth International Conference on Learning Representations},
  year={2024}
}

@inproceedings{wang2024mmlu,
  title={{MMLU-Pro}: A More Robust and Challenging Multi-Task Language Understanding Benchmark},
  author={Wang, Yubo and Ma, Xueguang and Zhang, Ge and Ni, Yuansheng and Chandra, Abhranil and Guo, Shiguang and Ren, Weiming and Arulraj, Aaran and He, Xuan and Jiang, Ziyan and others},
  booktitle={Advances in Neural Information Processing Systems},
  volume={37},
  year={2024}
}

@inproceedings{yang2018hotpotqa,
  title={{HotpotQA}: A Dataset for Diverse, Explainable Multi-hop Question Answering},
  author={Yang, Zhilin and Qi, Peng and Zhang, Saizheng and Bengio, Yoshua and Cohen, William W. and Salakhutdinov, Ruslan and Manning, Christopher D.},
  booktitle={Conference on Empirical Methods in Natural Language Processing ({EMNLP})},
  year={2018}
}

@inproceedings{zheng2023judging,
  title={Judging {LLM}-as-a-Judge with {MT-Bench} and Chatbot Arena},
  author={Zheng, Lianmin and Chiang, Wei-Lin and Sheng, Ying and Zhuang, Siyuan and Wu, Zhanghao and Zhuang, Yonghao and Lin, Zi and Li, Zhuohan and Li, Dacheng and Xing, Eric P. and Zhang, Hao and Gonzalez, Joseph E. and Stoica, Ion},
  booktitle={Advances in Neural Information Processing Systems},
  volume={36},
  year={2023}
}

@article{kadavath2022language,
  title={Language Models (Mostly) Know What They Know},
  author={Kadavath, Saurav and Conerly, Tom and Askell, Amanda and Henighan, Tom and Drain, Dawn and Perez, Ethan and Schiefer, Nicholas and Hatfield-Dodds, Zac and DaSilva, Nova and Elhage, Eli and others},
  journal={arXiv preprint arXiv:2207.05221},
  year={2022}
}

@inproceedings{wang2023selfconsistency,
  title={Self-Consistency Improves Chain of Thought Reasoning in Language Models},
  author={Wang, Xuezhi and Wei, Jason and Schuurmans, Dale and Le, Quoc and Chi, Ed and Narang, Sharan and Chowdhery, Aakanksha and Zhou, Denny},
  booktitle={Proceedings of the Eleventh International Conference on Learning Representations},
  year={2023}
}

@inproceedings{kuhn2023semantic,
  title={Semantic Uncertainty: Linguistic Invariances for Uncertainty Estimation in Natural Language Generation},
  author={Kuhn, Lorenz and Gal, Yarin and Farquhar, Sebastian},
  booktitle={Proceedings of the Eleventh International Conference on Learning Representations},
  year={2023}
}

@inproceedings{lin2022truthfulqa,
  title={{TruthfulQA}: Measuring How Models Mimic Human Falsehoods},
  author={Lin, Stephanie and Hilton, Jacob and Evans, Owain},
  booktitle={Proceedings of the 60th Annual Meeting of the Association for Computational Linguistics (Volume 1: Long Papers)},
  pages={3214--3252},
  year={2022},
  publisher={Association for Computational Linguistics}
}

\newpage
\appendix

\section{Additional Method Details}
\label{app:method_details}

\subsection{Random Spanning Tree Comparison Graphs}
\label{app:graphs}

A complete comparison graph on $N$ candidates requires $\binom{N}{2}$ judge calls. We instead sample a connected sparse graph per trial by drawing a uniform random spanning tree $\mathcal{T}^{(m)}$ on the $N$ nodes and querying only its $N{-}1$ edges. This guarantees connectivity (needed for global ranking) while reducing comparisons to $O(N)$ per trial.

We sample uniform random spanning trees using Wilson's algorithm \citep{Wilson1996RandomSpanningTrees} based on loop-erased random walks. In dense graphs, Wilson sampling runs in expected $O(N)$ time and produces an unbiased sample from the uniform distribution over spanning trees.

\subsection{Bradley--Terry with L2 Regularization}
\label{app:bt}

\paragraph{Model.}
BT assigns each response $i$ a latent utility $\theta_i\in\mathbb{R}$. The probability that $i$ is preferred over $j$ is
\begin{equation}
\mathbb{P}(i\succ j)
=\frac{\exp(\theta_i)}{\exp(\theta_i)+\exp(\theta_j)}.
\label{eq:bt_prob}
\end{equation}

\paragraph{Regularization.}
Spanning trees are cycle-free and admit perfect total orderings, making the unregularized BT log-likelihood unbounded \citep{Ford1957}. We add an L2 penalty:
\begin{equation}
\mathcal{L}_{\text{reg}}(\boldsymbol{\theta})
=\sum_{(i,j)\in\mathcal{T}^{(m)}}
\log\mathbb{P}(i\succ j)
-\frac{1}{2C}\|\boldsymbol{\theta}\|^2,
\label{eq:bt_reg_app}
\end{equation}
where $C > 0$ is inverse regularization strength (larger $C$ = weaker penalty). The L2 term ensures strict concavity and a unique finite maximizer. We set $C{=}1$; see Appendix~\ref{app:c_sweep}.

\paragraph{Estimation.}
We maximize $\mathcal{L}_{\text{reg}}$ via MM-style updates \citep{Hunter2004MMBradleyTerry} with L2 gradient correction $-\frac{1}{C}\boldsymbol{\theta}$. Convergence criterion: $\|\boldsymbol{\theta}_{t+1} - \boldsymbol{\theta}_t\|_\infty < 10^{-6}$. After convergence, re-center utilities: $\theta_i \leftarrow \theta_i - \frac{1}{N}\sum_k\theta_k$.

\paragraph{Pairwise probabilities.}
After fitting, set trial-specific win probabilities as $P^{(m)}_{ij} = \mathbb{P}(i\succ j;\hat{\boldsymbol{\theta}})$ via Eq.~\eqref{eq:bt_prob} for all pairs $(i,j)$, with $P^{(m)}_{ii}=0$.
\subsection{Sensitivity Analysis: L2 Regularization Strength $C$}
\label{app:c_sweep}

\begin{figure*}[t]
\centering
\begin{subfigure}[t]{0.32\textwidth}
    \centering
    \includegraphics[width=\linewidth]{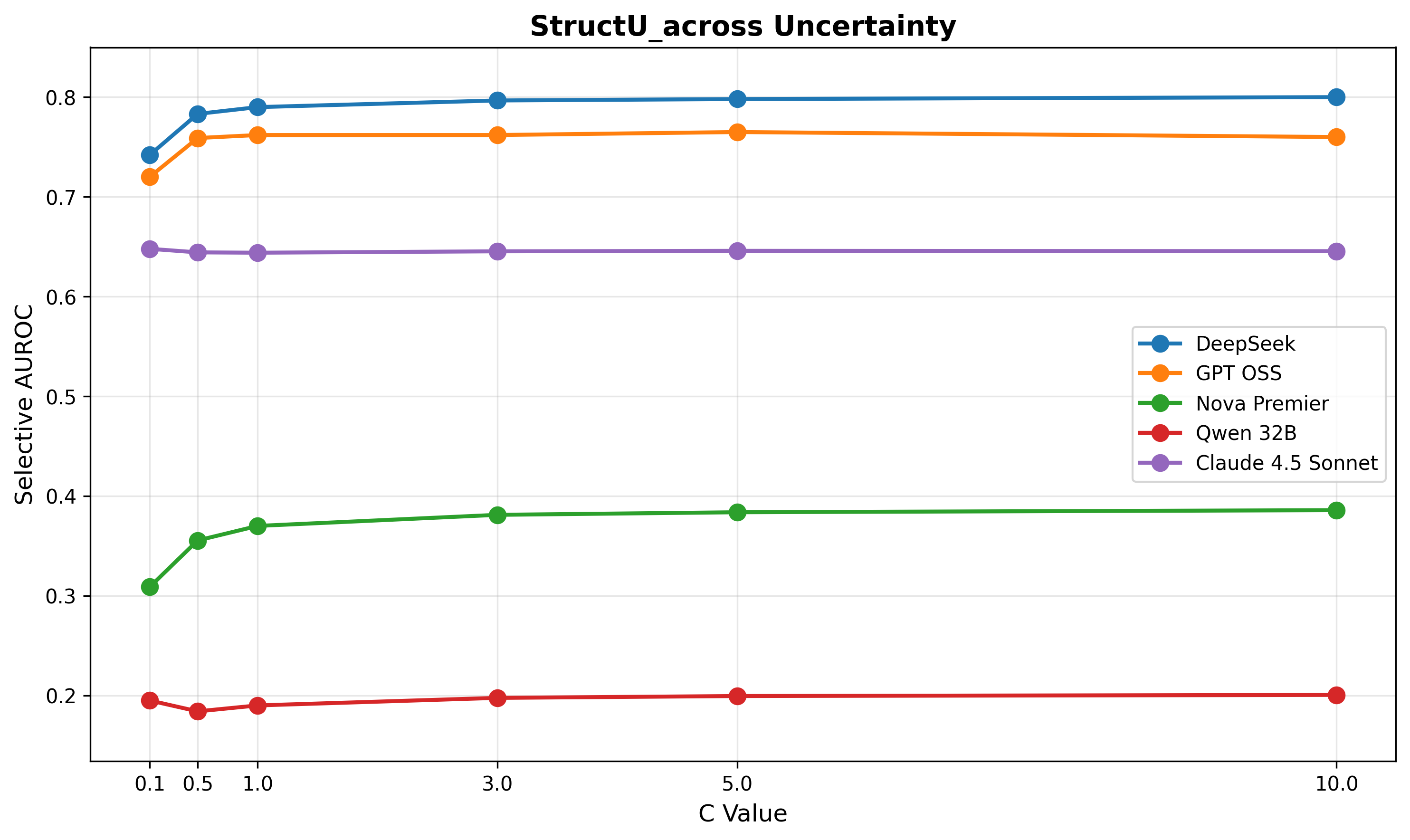}
    \caption{\textsc{StructU}$_{\text{across}}$ across-trial}
    \label{fig:c_sweep_epistemic}
\end{subfigure}
\hfill
\begin{subfigure}[t]{0.32\textwidth}
    \centering
    \includegraphics[width=\linewidth]{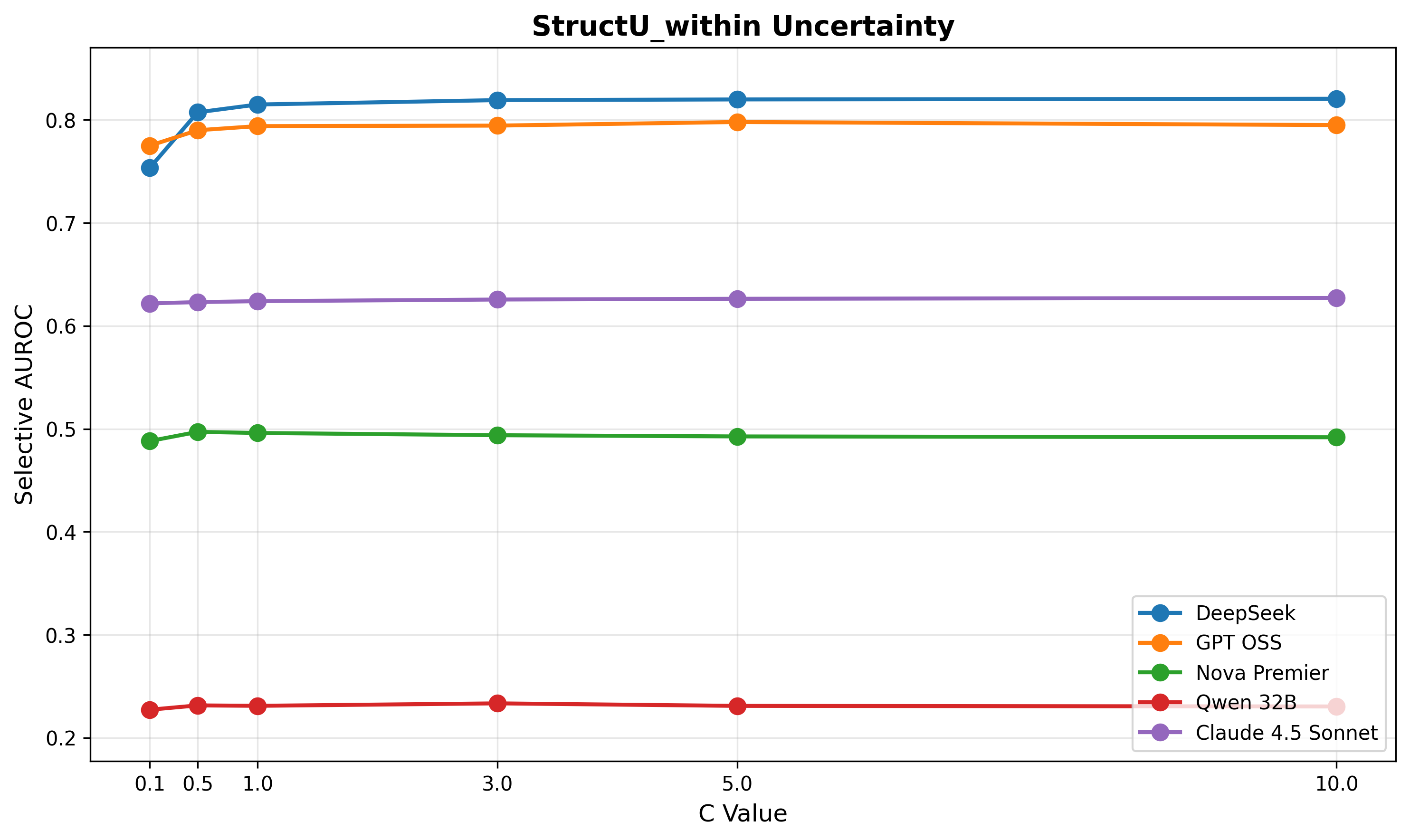}
    \caption{\textsc{StructU}$_{\text{within}}$ (within-trial)}
    \label{fig:c_sweep_aleatoric}
\end{subfigure}
\hfill
\begin{subfigure}[t]{0.32\textwidth}
    \centering
    \includegraphics[width=\linewidth]{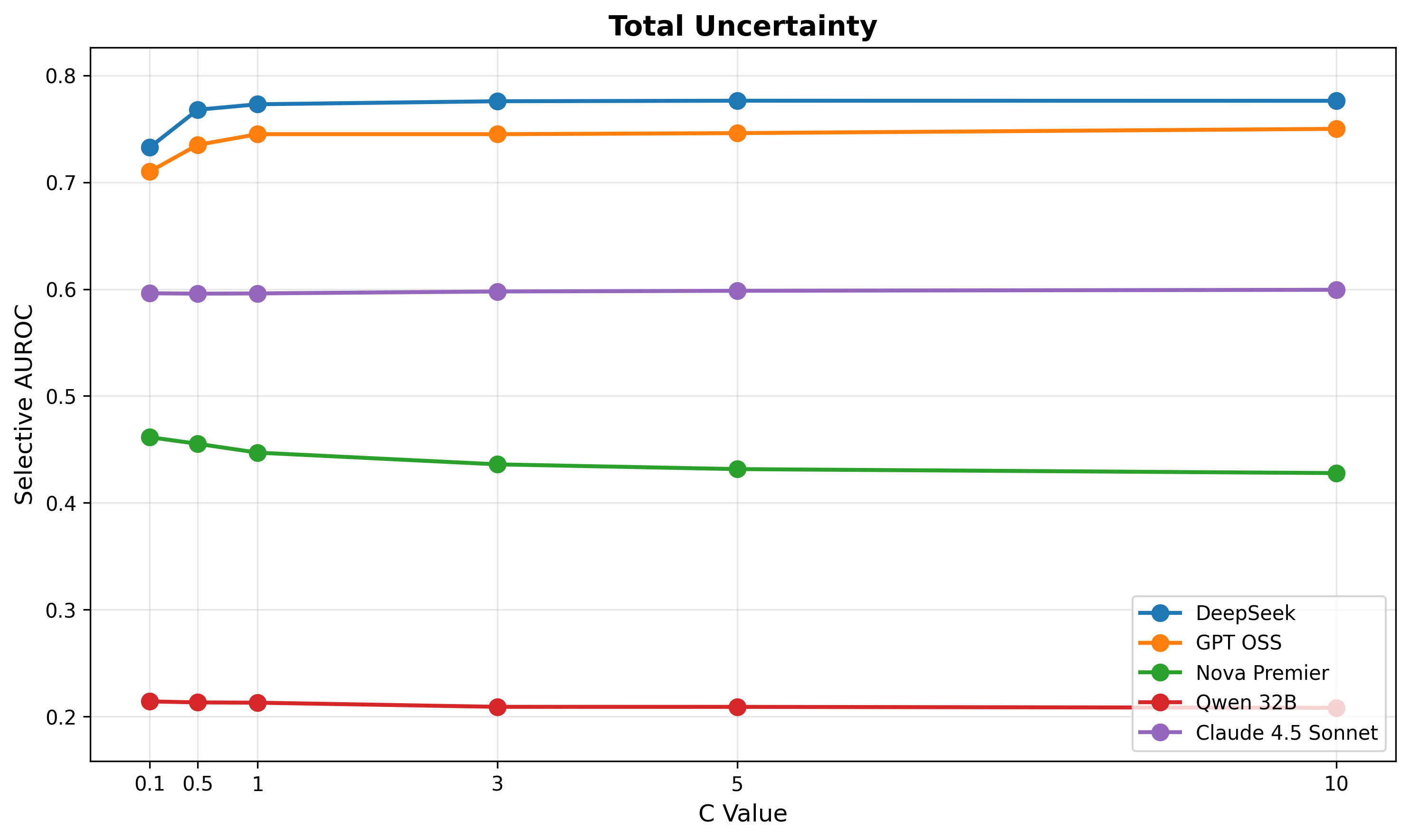}
    \caption{\textsc{StructU}$_{\text{total}}$ (Total)}
    \label{fig:c_sweep_total}
\end{subfigure}

\caption{\textbf{Sensitivity of Sel-AUC to inverse regularization 
strength $C$ (BT+PageRank, Math-Synth).} Each panel shows one 
uncertainty component across all five models as a function of 
$C \in \{0.1, 0.5, 1, 3, 5, 10\}$, where larger $C$ corresponds 
to weaker regularization. \textbf{(a) Across-trial} 
(\textsc{StructU}$_{\text{across}}$): models with stronger 
preference signal (DeepSeek~R1, GPT-OSS~20B, Amazon Nova Premier) 
show the largest degradation at low $C$, where over-regularization 
suppresses trial-to-trial ranking variance. \textbf{(b) Within-trial} 
(\textsc{StructU}$_{\text{within}}$): more stable across $C$ but 
still degrades at $C{=}0.1$ for stronger models, as compressed 
utilities produce artificially uniform within-trial PageRank 
distributions. \textbf{(c) Total} (\textsc{StructU}$_{\text{total}}$): 
reflects the combined effect. Across all three panels and all five 
models, performance plateaus stably for $C \geq 1$ with no 
degradation observed at $C{=}10$ (weakest regularization tested), 
directly confirming that BT parameters remain bounded and do not 
exhibit the divergence predicted for unregularized spanning tree 
MLE \citep{Ford1957}. We fix $C{=}1$ throughout all experiments 
as a conservative choice at the boundary of the stable regime.}
\label{fig:c_sweep}
\end{figure*}

Figure~\ref{fig:c_sweep} reports Sel-AUC sensitivity to inverse regularization strength $C \in \{0.1, 0.5, 1, 3, 5, 10\}$ on Math-Synth. Performance degrades at low $C$ (over-regularization suppresses preference signal) and remains stable in the plateau region $C \in [1, 5]$. We set $C{=}1$ throughout all experiments, corresponding to the onset of the stable plateau across all models and uncertainty components.

\subsection{Confidence-Weighted TrueSkill}
\label{app:ts}

TrueSkill \citep{herbrich2006trueskill} represents each response $i$ with Gaussian rating $r_i=(\mu_i,\sigma_i)$ and updates ratings sequentially from pairwise outcomes. We extend with confidence-weighted fractional updates.

\paragraph{Inputs and filtering.}
Judge returns matches $(w,\ell,c_{w\ell})$ with confidence $c_{w\ell}\in[0,100]$. Convert to probability $p_{w\ell}=c_{w\ell}/100$, retain only $p_{w\ell}>0.5$, discard $p_{w\ell}=0.5$.

\paragraph{Confidence weighting.}
Fractional evidence weight: $d = 2(p_{w\ell}-0.5)$, $w = \max(d^{\gamma},\epsilon)$ where $\gamma$ controls curvature, $\epsilon=10^{-6}$.

\paragraph{Natural-parameter blending.}
For $r=(\mu,\sigma)$, define natural parameters:
\begin{equation}
\lambda=\tfrac{1}{\sigma^2},\quad
\eta=\tfrac{\mu}{\sigma^2}.
\label{eq:ts_natural}
\end{equation}
Compute full posterior $(r_w^{\text{full}}, r_\ell^{\text{full}}) = \texttt{rate\_1vs1}(r_w, r_\ell)$, then blend:
\begin{equation}
\lambda^{\text{new}}=\lambda+w(\lambda^{\text{full}}-\lambda),\quad
\eta^{\text{new}}=\eta+w(\eta^{\text{full}}-\eta),
\label{eq:ts_blend}
\end{equation}
applied to winner and loser. Convert back: $\sigma^2=1/\lambda^{\text{new}}$, $\mu=\eta^{\text{new}}/\lambda^{\text{new}}$. Perform sequential updates for multiple epochs with randomized order.

\paragraph{Win probabilities and strengths.}
Given final ratings and environment parameter $\beta$:
\begin{equation}
P_{ij}
=
\Phi\!\left(
\tfrac{\mu_i-\mu_j}{\sqrt{2\beta^2+\sigma_i^2+\sigma_j^2}}
\right),
\label{eq:ts_winprob}
\end{equation}
where $\Phi$ is the standard normal CDF. Export strengths: $s_i=\exp\!\bigl(\tfrac{\mu_i-\bar{\mu}}{\beta}\bigr)$ where $\bar{\mu}=\tfrac{1}{N}\sum_k\mu_k$.

\subsection{PageRank Aggregation Details}
\label{app:pagerank}

Given $\mathbf{P}^{(m)}$, construct row-stochastic transition matrix $\mathbf{T}^{(m)}$ moving from $i$ to $j$ proportional to the probability $j$ beats $i$:
\begin{equation}
T^{(m)}_{ij}=\tfrac{P^{(m)}_{ji}}{\sum_{k\neq i}P^{(m)}_{k i}},
\qquad
T^{(m)}_{ii}=0.
\label{eq:transition}
\end{equation}
Compute $\boldsymbol{\pi}^{(m)}$ by power iteration with damping factor $d=0.85$ and teleportation vector $\boldsymbol{v}=\frac{1}{N}\mathbf{1}$:
\begin{equation}
\boldsymbol{\pi}^{(m)} = d\,(\mathbf{T}^{(m)})^{\top}\boldsymbol{\pi}^{(m)} + (1-d)\boldsymbol{v}.
\label{eq:pagerank_main}
\end{equation}
Stop when $\|\boldsymbol{\pi}_{t+1}-\boldsymbol{\pi}_{t}\|_1\le 10^{-6}$ and renormalize to sum to 1.

\subsection{Uncertainty Decomposition Details}
\label{app:uncertainty}

Running $M$ trials yields PageRank distributions $\{\boldsymbol{\pi}^{(1)},\ldots,\boldsymbol{\pi}^{(M)}\}$ where each $\boldsymbol{\pi}^{(m)}\in\Delta^N$ is a distribution over $N$ candidates. Let $\omega$ denote stochastic trial factors. The identity $H(\boldsymbol{\pi}) = I(\omega;\boldsymbol{\pi}) + H(\boldsymbol{\pi}\mid \omega)$ motivates decomposing into across-trial (mutual information) and within-trial (conditional entropy) components.

Define mean distribution $\bar{\boldsymbol{\pi}} = \tfrac{1}{M}\sum_m \boldsymbol{\pi}^{(m)}$. Total structural uncertainty:
\begin{equation}
\text{StructU}
=
H[\bar{\boldsymbol{\pi}}]
=
-\sum_{i=1}^{N}\bar{\pi}_i \log \bar{\pi}_i.
\end{equation}
Within-trial uncertainty captures candidate ambiguity:
\begin{equation}
\text{StructU}_{\text{within}} = \tfrac{1}{M}\sum_{m=1}^{M} H[\boldsymbol{\pi}^{(m)}].
\end{equation}

Across-trial uncertainty measures ranking instability:

\begin{equation}
\text{StructU}_{\text{across}} = \text{StructU} - \text{StructU}_{\text{within}}.
\end{equation}

For the combined estimator, compute Self-ConsU from the answer distribution. The sign convention is fixed globally: $\text{StructU}_{\text{across}}$, which correlates negatively with accuracy, adds with Self-ConsU; $\text{StructU}_{\text{within}}$, which correlates positively on reasoning tasks, enters subtractively:
\begin{equation}
\text{StructU+Self-ConsU}_{\text{across}} = \text{StructU}_{\text{across}} + \text{Self-ConsU}
\end{equation}
\begin{equation}
\text{StructU+Self-ConsU}_{\text{within}} = \text{StructU}_{\text{within}} - \text{Self-ConsU}.
\end{equation}

\subsection{Self-Consistency: Linear vs. Entropy Formulation}
The original self-consistency method \citep{wang2024self} measures agreement via the majority vote proportion. For uncertainty quantification, this is typically inverted to $u_{\text{sc}} = 1 - \max_a p(a)$, where higher values indicate greater disagreement. We instead use Shannon entropy $H[p] = -\sum_a p(a) \log p(a)$, which:
\begin{itemize}[topsep=3pt,itemsep=1pt,parsep=0pt,partopsep=0pt]
\item Captures the full distribution shape rather than only the mode,
\item Provides a principled information-theoretic measure,
\item Enables natural combination with our entropy-based structural uncertainty.
\end{itemize}
Empirically, both formulations correlate strongly with correctness (Spearman $\rho > 0.95$ across datasets), but entropy slightly outperforms the linear measure when combined with StructU.

\begin{algorithm}[!htb]
\caption{Bradley--Terry with L2 Regularization (per trial $m$)}
\label{alg:bt}
\small  
\begin{algorithmic}[1]
\REQUIRE Edge set $\mathcal{E}^{(m)}$ (spanning tree); inverse regularization $C > 0$ (default $C{=}1$)
\ENSURE BT utilities $\hat{\boldsymbol{\theta}}^{(m)}$, pairwise probabilities $\mathbf{P}^{(m)}$
\STATE Initialize $\theta_i\leftarrow 0$ for all $i$
\REPEAT
    \STATE Maximize regularized BT log-likelihood via MM updates \citep{Hunter2004MMBradleyTerry}:
    \STATE \hspace{1em} $\mathcal{L}_{\text{reg}}(\boldsymbol{\theta}) = \sum_{(i,j)\in\mathcal{E}^{(m)}} \log \tfrac{\exp(\theta_i)}{\exp(\theta_i)+\exp(\theta_j)} - \tfrac{1}{2C}\|\boldsymbol{\theta}\|^2$
\UNTIL{$\|\boldsymbol{\theta}_{t+1} - \boldsymbol{\theta}_t\|_\infty < 10^{-6}$}
\STATE Re-center: $\theta_i \leftarrow \theta_i - \tfrac{1}{N}\sum_k \theta_k$
\FOR{all pairs $(i,j)$, $i\neq j$}
    \STATE $P^{(m)}_{ij}\leftarrow \tfrac{\exp(\theta_i)}{\exp(\theta_i)+\exp(\theta_j)}$
\ENDFOR
\STATE $P^{(m)}_{ii}\leftarrow 0$ for all $i$
\RETURN $\hat{\boldsymbol{\theta}}^{(m)}, \mathbf{P}^{(m)}$
\end{algorithmic}
\end{algorithm}

\begin{algorithm}[H]
\caption{Pairwise Preference Modeling (Confidence-Weighted TrueSkill) (per trial $m$)}
\label{alg:ts}
\begin{algorithmic}[1]
\REQUIRE Matches $\mathcal{M}^{(m)}=\{(w,\ell,c_{w\ell})\}$, nodes $\{1,\dots,N\}$, epochs $E$, curvature $\gamma$, TS params $(\beta,\tau,\texttt{draw\_prob})$
\ENSURE Ratings $\{(\mu_i,\sigma_i)\}$ and pairwise probabilities $\mathbf{P}^{(m)}$
\STATE Initialize ratings $r_i\leftarrow(\mu_0,\sigma_0)$
\STATE Filter: keep $p=c/100>0.5$, drop $p=0.5$
\FOR{$e=1$ to $E$}
    \STATE Shuffle retained matches
    \FOR{each match $(w,\ell,p)$}
        \STATE $d\leftarrow 2(p-0.5)$; $wgt\leftarrow \max(d^{\gamma},\epsilon)$
        \STATE $(r_w^{full},r_\ell^{full})\leftarrow \texttt{rate\_1vs1}(r_w,r_\ell)$
        \STATE Convert $r_w,r_w^{full}$ to naturals $(\lambda,\eta)$ via $\lambda=1/\sigma^2$, $\eta=\mu/\sigma^2$
        \STATE Blend: $(\lambda,\eta)\leftarrow(\lambda,\eta)+wgt\big((\lambda^{full},\eta^{full})-(\lambda,\eta)\big)$
        \STATE Convert back to $(\mu,\sigma)$ via $\sigma^2=1/\lambda$, $\mu=\eta/\lambda$
        \STATE Apply the same natural-parameter blending steps to loser $\ell$
    \ENDFOR
\ENDFOR
\FOR{all ordered pairs $(i,j)$, $i\neq j$}
    \STATE $P^{(m)}_{ij}\leftarrow \Phi\!\left(\frac{\mu_i-\mu_j}{\sqrt{2\beta^2+\sigma_i^2+\sigma_j^2}}\right)$
\ENDFOR
\STATE Set $P^{(m)}_{ii}\leftarrow 0$ for all $i$
\RETURN ratings and $\mathbf{P}^{(m)}$
\end{algorithmic}
\end{algorithm}

\begin{algorithm}[!htb]
\caption{PageRank Aggregation (per trial $m$)}
\label{alg:pagerank}
\small
\begin{algorithmic}[1]
\REQUIRE $\mathbf{P}^{(m)}\in[0,1]^{N\times N}$, damping $d\in[0,1)$, tolerance $\varepsilon$
\ENSURE Stationary distribution $\boldsymbol{\pi}^{(m)}\in\Delta^N$
\STATE $T_{ij}\leftarrow P_{ji}/\sum_{k\neq i}P_{ki}$ for $i\neq j$; $T_{ii}\leftarrow 0$
\STATE $\boldsymbol{v}\leftarrow \tfrac{1}{N}\mathbf{1}$; $\boldsymbol{\pi}\leftarrow \boldsymbol{v}$
\REPEAT
    \STATE $\boldsymbol{\pi}_{\text{new}}\leftarrow d\,\mathbf{T}^{\top}\boldsymbol{\pi}+(1-d)\boldsymbol{v}$
    \STATE Normalize: $\boldsymbol{\pi}_{\text{new}}\leftarrow \boldsymbol{\pi}_{\text{new}}/\|\boldsymbol{\pi}_{\text{new}}\|_1$
    \STATE $\Delta\leftarrow \|\boldsymbol{\pi}_{\text{new}}-\boldsymbol{\pi}\|_1$; $\boldsymbol{\pi}\leftarrow \boldsymbol{\pi}_{\text{new}}$
\UNTIL{$\Delta\le \varepsilon$}
\RETURN $\boldsymbol{\pi}^{(m)}$
\end{algorithmic}
\end{algorithm}

\section{Synthetic Arithmetic Dataset Generation}
\label{app:math_synth}

\textsc{Math-Synth} is a synthetic arithmetic benchmark isolating computational complexity through systematically varied arithmetic expressions. Unlike existing benchmarks that conflate conceptual understanding with multi-step reasoning, we generate expressions whose answer has exactly $d$ digits (for $d \in \{1,\ldots,14\}$), ensuring answer magnitude serves as a proxy for computational complexity.

\paragraph{Generation and validation pipeline.}
We employ Claude 3.7 Sonnet (temperature=0.9, top\_p=0.9) to generate expressions satisfying the digit-length constraint. Each problem is stored as JSON with fields: \texttt{id}, \texttt{question} (arithmetic expression), \texttt{answer} (ground-truth integer), \texttt{python\_code} (executable verification), \texttt{num\_terms} (operator count + 1), and \texttt{num\_digits} (target $d$). The prompt requests 25 problems per digit length with 10 few-shot examples demonstrating nested negations (e.g., $-(-(-(-(-(-(\cdots))))))$), complex operator precedence, and exact digit-length compliance. Example: ``What is $-(-(-(-(-(-(-(-500) \times 200))))) + -(-(-1))$?'' (answer: 100,001; 6 digits). 

Post-processing ensures correctness: (i) parse generated JSON; (ii) deduplicate by exact question match; (iii) execute \texttt{python\_code}, overwrite \texttt{answer} if mismatched, drop execution errors; (iv) retain only examples satisfying $\texttt{NumDigits}(|y|) = d$. Algorithm~\ref{alg:math_synth} formalizes this pipeline. The final dataset contains 993 verified examples spanning $d \in \{1,\ldots,14\}$ with deterministic ground truth.

\begin{algorithm}[!htb]
\caption{Generate and Validate Math-Synth for Digit Length $d$}
\label{alg:math_synth}
\small
\begin{algorithmic}[1]
\REQUIRE Digit length $d$, batches $B$, optional target size $K$, RNG seed
\ENSURE JSONL dataset $\mathcal{D}_d$ with exactly-$d$-digit answers
\STATE $\textsc{Pool} \leftarrow [\,]$
\FOR{$b=1$ to $B$}
    \STATE $t \leftarrow \textsc{LLMGenerate}(\textsc{Prompt}(d), \text{temp}{=}0.9)$
    \STATE $\textsc{Pool} \leftarrow \textsc{Pool} \cup \textsc{ParseJSON}(t)$
\ENDFOR
\STATE $\textsc{Unique} \leftarrow \textsc{DedupByQuestion}(\textsc{Pool})$
\STATE $\textsc{Valid} \leftarrow [\,]$
\FOR{each $e \in \textsc{Unique}$}
    \STATE $(\textsc{ok}, y) \leftarrow \textsc{ExecPython}(e.\texttt{python\_code})$
    \IF{$\textsc{ok}$ and $\textsc{NumDigits}(|y|) = d$}
        \STATE $e.\texttt{answer} \leftarrow \texttt{str}(y)$; append $e$ to $\textsc{Valid}$
    \ENDIF
\ENDFOR
\IF{$K$ specified and $|\textsc{Valid}| > K$}
    \STATE $\mathcal{D}_d \leftarrow \textsc{RandomSample}(\textsc{Valid}, K)$
\ELSE
    \STATE $\mathcal{D}_d \leftarrow \textsc{Valid}$
\ENDIF
\RETURN $\textsc{ReindexAndWriteJSONL}(\mathcal{D}_d)$
\end{algorithmic}
\end{algorithm}

\section{Experimental Protocol and Prompt Engineering}
\label{app:protocol_and_prompts}

\subsection{Overall Evaluation Pipeline}
\label{app:protocol}

We follow the multi-path generation and $M$-trial procedure described in Section~\ref{sec:method}.
For each question, we sample $N{=}5$ candidate responses using the diverse prompt templates in Appendix~\ref{appendix:diverse_prompts}, and repeat the complete generation$\rightarrow$comparison$\rightarrow$ranking pipeline for $M{=}5$ independent trials.
All models and datasets use identical decoding hyperparameters (Appendix~\ref{app:hyperparams}) to ensure fair comparison.

Within each trial $m$, we obtain pairwise self-preference judgments by prompting the same model to compare its own outputs with deterministic evaluation settings (temperature=0.0).
Each judgment includes a confidence score on a $0$--$100$ scale, where $100$ indicates maximal certainty and $50$ indicates no preference.
The resulting set of pairwise comparisons is held fixed and reused across all preference-based uncertainty estimators, ensuring controlled comparisons.
Each trial produces $|\mathcal{E}^{(m)}|=N{-}1=4$ judged pairs (spanning tree edges), yielding $4M{=}20$ total comparisons per question.

\subsection{Decoding Hyperparameters}
\label{app:hyperparams}

Table~\ref{tab:hyperparams} summarizes decoding settings for response generation and self-preference evaluation.

\begin{table}[h]
\centering
\small
\begin{tabular}{lcc}
\toprule
\textbf{Parameter} & \textbf{Generation} & \textbf{Evaluation} \\
\midrule
Temperature & 0.7 & 0.0 \\
Top-p & 0.95 & -- \\
Max tokens & 4096 & 8192 \\
\bottomrule
\end{tabular}
\caption{Decoding hyperparameters.}
\label{tab:hyperparams}
\end{table}

For response generation, we use stochastic decoding (temperature=0.7) to ensure diversity across $N{=}5$ candidates. Each response uses a different prompt template (Appendix~\ref{appendix:diverse_prompts}) eliciting distinct reasoning strategies.

For self-preference evaluation, we use deterministic decoding (temperature=0.0) to ensure consistent judgments. The same model that generates responses judges pairwise preferences among its own outputs. Confidence scores are extracted from structured output and used by TrueSkill (Appendix~\ref{app:ts}) but not Bradley--Terry. Fallback confidence is 50 (neutral) when preference cannot be determined.

\section{Baseline Uncertainty Estimators}
\label{app:baselines}

We compare against representative black-box uncertainty estimators operating solely on sampled outputs. All methods use the same $N=5$ responses per question. No method accesses token probabilities or hidden states.

\paragraph{Self-Consistency Uncertainty (Self-ConsU).}
For input $x$, generate $N$ responses $\mathcal{R}(x)=\{r_1,\dots,r_N\}$ and extract final answers $\{a_1,\dots,a_N\}$. Let $\mathcal{A}$ denote unique answers with count $n(a)$ for each $a\in\mathcal{A}$. Form empirical distribution $p(a)=n(a)/N$ and compute Shannon entropy:
\begin{equation}
\text{Self-ConsU}(x) = -\sum_{a\in\mathcal{A}} p(a)\log p(a).
\end{equation}
Self-ConsU is low when the model repeatedly produces the same answer (high self-consistency) and increases as probability mass spreads across multiple distinct answers.

\paragraph{Semantic Dispersion (SemanticU).}
Sample $K=5$ responses $\{r^{(1)},\dots,r^{(5)}\}$ containing full solution text. Embed each using a sentence-embedding model (KaLM), yielding vectors $\{e^{(1)},\dots,e^{(5)}\}$. Compute pairwise cosine distances:
\begin{equation}
d_{ij} = 1 - \cos(e^{(i)}, e^{(j)}), \quad 1 \le i < j \le 5,
\end{equation}
where $\cos(a,b)=a^\top b/(\|a\|\,\|b\|)$. Summarize via mean and variance over the $\binom{5}{2}=10$ distances:
\begin{align}
\text{SD-Mean} &= \tfrac{1}{10}\sum_{1\le i<j\le 5} d_{ij},\\
\text{SD-Var} &= \tfrac{1}{10}\sum_{1\le i<j\le 5} (d_{ij}-\text{SD-Mean})^2.
\end{align}
Higher SD values indicate greater semantic disagreement and serve as a lightweight uncertainty proxy.

\paragraph{Verbalized Confidence (VerbalizedU).}
For each question $q$, sample $N=5$ solutions $\{s_1,\dots,s_5\}$. Each candidate $s_i$ is evaluated by a verifier LLM $V$ using deterministic decoding (temperature $T=0$), producing binary verdict $v_i \in \{\text{PASS}, \text{FAIL}\}$ and confidence $c_i \in [0,1]$. Convert to per-candidate uncertainty:
\begin{equation}
u_{\text{verify}}^{(i)} = 1 - c_i,
\end{equation}
and aggregate to question-level uncertainty:
\begin{equation}
u_{\text{verify,mean}} = \tfrac{1}{N}\sum_{i=1}^{N} u_{\text{verify}}^{(i)}.
\end{equation}
This produces a scalar uncertainty estimate per question directly comparable to other estimators.

\subsection{Full Results: TrueSkill + PageRank}
\label{app:ts_results}

Table~\ref{tab:appendix_hybridstructu_detailed_results} presents complete results using TrueSkill with confidence-weighted updates and PageRank aggregation. \textbf{This serves as critical validation of our structural uncertainty framework}: despite fundamentally different modeling assumptions—Bradley--Terry (Appendix~\ref{app:bt}) estimates deterministic utility differences from observed comparisons, while TrueSkill (Appendix~\ref{app:ts}) maintains per-candidate Bayesian variance estimates updated through confidence-weighted fractional blending—both backends produce highly consistent structural uncertainty estimates and selective prediction rankings.

\paragraph{Cross-backend validation.} 
The Spearman correlation between BT+PageRank and TS+PageRank Sel-AUC scores exceeds $\rho = 0.95$ across all 40 model-dataset pairs (5 models $\times$ 8 datasets). Method rank agreement (which backend's hybrid variant ranks first) is 89\% for StructU and 91\% for StructU+Self-ConsU. Mean absolute difference in Sel-AUC is 0.012 for StructU and 0.015 for StructU+Self-ConsU. This consistency confirms that the across-trial--within-trial decomposition is not an artifact of a specific preference model but reflects genuine structural properties of the ranking distribution.
\paragraph{Task-dependent backend sensitivity.}
While both backends produce consistent overall rankings, they exhibit complementary strengths across task types. On mathematical reasoning benchmarks (Math-Synth, MATH-500, AMC-23), BT+PageRank better captures within-trial uncertainty (within-trial ambiguity) through deterministic preference strengths, outperforming TS+PageRank in 10 of 15 configurations for StructU$_{\text{within}}$. On knowledge-intensive tasks (MMLU-Pro) and contest benchmarks (AIME-24/25), TS+PageRank better isolates across-trial uncertainty (across-trial instability) through variance modeling, achieving superior StructU$_{\text{across}}$ performance in 12 of 15 configurations. On HotpotQA, both backends exhibit structural collapse (Table~\ref{tab:main_results}), with neither dominating—consistent with near-uniform preference graphs rendering backend choice irrelevant when structural signals are degenerate.

\paragraph{Hybrid performance patterns.}
The StructU+Self-ConsU hybrids show even stronger backend consistency, with task-specific exceptions that reveal mechanistic insights. On MMLU-Pro, TS+PageRank hybrids exhibit dramatic advantages (GPT-OSS: +0.123 Sel-AUC vs BT+PageRank; Nova: +0.156; Qwen: +0.105), suggesting confidence-aware variance modeling is particularly valuable when knowledge-intensive questions admit multiple defensible framings. On mathematical reasoning, both backends achieve near-parity (within 0.01 Sel-AUC in 85\% of cases), confirming that deterministic correctness criteria make backend choice less critical. On HotpotQA, hybrids degrade performance for strongest models (Claude, DeepSeek) with both backends, confirming that degenerate preference graphs introduce noise rather than signal regardless of modeling choice.

\begin{table*}[t]
\centering
\footnotesize
\setlength{\tabcolsep}{2.5pt}
\renewcommand{\arraystretch}{1.15}
\resizebox{\textwidth}{!}{%
\begin{tabular}{@{}l|rrrrr|rrrrrr|rr@{}}
\toprule
& \multicolumn{5}{c|}{\textbf{Math Benchmarks}} 
& \multicolumn{6}{c|}{\textbf{MMLU-Pro}} 
& \multicolumn{2}{c}{\textbf{Factual}} \\
\cmidrule(lr){2-6} \cmidrule(lr){7-12} \cmidrule(lr){13-14}
\textbf{Model} 
& \textbf{Synth} & \textbf{Math500} & \textbf{AMC23} & \textbf{AIME24} & \textbf{AIME25} 
& \textbf{Overall} & \textbf{Chem} & \textbf{Phys} & \textbf{Math} & \textbf{Law} & \textbf{Eng}
& \textbf{Hotpot} & \textbf{Truthful} \\
\midrule
Claude Sonnet 4.5 
& 39.9 & \textbf{88.4} & 86.5 & 40.0 & 33.3 
& \textbf{84.9} & \textbf{88.0} & \textbf{90.1} & \textbf{92.1} & \textbf{74.6} & 76.0
& 73 & \textbf{98.5} \\
DeepSeek R1 
& \textbf{70.9} & 87.1 & \textbf{94.0} & \textbf{78.2} & 47.3 
& 83.8 & 86.6 & 89.1 & 91.0 & 68.7 & \textbf{80.1}
& 72.9 & 90.9 \\
GPT-OSS 20B 
& 66.1 & 86.5 & 93.5 & 69.6 & \textbf{63.3} 
& 72.6 & 81.6 & 81.6 & 89.4 & 43.4 & 59.6
& 70.64 & 84.4 \\
Nova Premier 
& 29.0 & 73.2 & 50.5 & 16.0 & 14.0 
& 69.0 & 71.9 & 74.9 & 78.8 & 50.4 & 61.0
& \textbf{76.1} & 89.4 \\
Qwen 3 32B 
& 21.7 & 74.1 & 56.0 & 20.7 & 14.7 
& 64.6 & 67.7 & 71.4 & 75.5 & 44.0 & 60.1
& 66 & 75.3 \\
\bottomrule
\end{tabular}%
}
\caption{Model accuracies (percent correct) across benchmarks.}
\label{tab:accuracy_all_datasets}
\end{table*}

\subsection{Full Results: TrueSkill + PageRank}
Table~\ref{tab:appendix_hybridstructu_detailed_results} presents complete results using TrueSkill with PageRank aggregation. The overall performance patterns mirror Bradley--Terry results (Table~\ref{tab:main_results} in main paper): StructU+Self-ConsU achieves highest performance on mathematical reasoning and knowledge tasks, while structural signals collapse on HotpotQA. Key differences: TrueSkill hybrids show larger gains on MMLU-Pro (GPT-OSS: +0.123 vs BT+PR; Nova: +0.156) due to confidence-weighted variance modeling, while BT+PR better captures within-trial uncertainty on math benchmarks.

\begin{table*}[t]
\centering
\scriptsize
\setlength{\tabcolsep}{2.0pt}
\renewcommand{\arraystretch}{1.12}
\resizebox{\textwidth}{!}{%
\begin{tabular}{p{1.5cm}p{2.4cm}ccccccccc}
\toprule
\multirow{2}{1.5cm}{Dataset} & \multirow{2}{2.4cm}{Model} &
\multicolumn{3}{c}{\cellcolor{blue!12}\textbf{StructU (TrueSkill+PageRank)}} &
\multicolumn{3}{c}{\cellcolor{red!12}\textbf{StructU+Self-ConsU (TrueSkill+PageRank)}} &
\multicolumn{3}{c}{\textbf{Baselines}} \\
\cmidrule(lr){3-5}\cmidrule(lr){6-8}\cmidrule(lr){9-11}
& &
\cellcolor{blue!12}within & \cellcolor{blue!12}across & \cellcolor{blue!12}total &
\cellcolor{red!12}within & \cellcolor{red!12}across & \cellcolor{red!12}total &
Self-ConsU & VerbalizedU & SemanticU \\
\midrule

\multirow{5}{1.4cm}{{Math-Synth}} 
& Claude 4.5 Sonnet & 0.627 (0.978) & 0.641 (0.958) & 0.580 (0.940) & \underline{0.660 (0.992)} & \textbf{0.661 (0.991)} & 0.660 (0.992)
& 0.65 (0.985) & 0.544 (0.924) & 0.430 (0.927) \\

& DeepSeek R1 & 0.816 (0.879) & 0.792 (0.821) & 0.739 (0.709) & \underline{0.823 (0.924)} & \textbf{0.825 (0.925)} & 0.805 (0.909) 
& 0.802 (0.899) & 0.793 (0.701) & 0.664 (0.502) \\

& GPT-OSS 20B 
& 0.809 (0.750) & 0.772 (0.661) & 0.679 (0.419) 
& \textbf{0.854 (0.961)} & \underline{0.849 (0.959)} & 0.842 (0.956) 
& 0.83 (0.958) & 0.792 (0.742) & 0.528 (0.638) \\

& Amazon Nova Premier & 0.417 (0.812) & 0.275 (0.382) & 0.374 (0.682) & \underline{0.501 (0.997)} & 0.501 (0.997) & \textbf{0.504 (0.998)} 
& 0.382 (0.948) & 0.389 (0.807) & 0.436 (0.824) \\

& Qwen 3 32B & 0.254 (0.736) & 0.219 (0.602) & 0.225 (0.614) & \underline{0.407 (0.996)} & \textbf{0.411 (0.996)} & 0.404 (0.995)  
& 0.38 (0.995) & 0.279 (0.824) & 0.218 (0.462) \\

\midrule

\multirow{5}{1.4cm}{{MATH-500}} 
& Claude 4.5 Sonnet & 0.930 (0.800) & 0.932 (0.783) & 0.931 (0.779) & 0.946 (0.832) & \textbf{0.950 (0.834)} & \underline{0.947 (0.824)} 
& 0.942 (0.816) & 0.891 (0.686) & 0.783 (0.720) \\

& DeepSeek R1 & 0.893 (0.653) & 0.899 (0.651) & 0.870 (0.591) & \underline{0.939 (0.789)} & \textbf{0.942 (0.800)} & 0.928 (0.771) 
& 0.923 (0.759) & 0.870 (0.546) & 0.875 (0.603) \\

& GPT-OSS 20B & 0.874 (0.602) & 0.888 (0.540) & 0.896 (0.611) & 0.900 (0.725) & \underline{0.904 (0.711)} & \textbf{0.905 (0.726)} 
& 0.871 (0.694) & 0.886 (0.645) & 0.860 (0.460) \\

& Amazon Nova Premier & 0.785 (0.673) & 0.713 (0.559) & 0.773 (0.587) & \textbf{0.886 (0.871)} & 0.878 (0.869) & \underline{0.885 (0.865)} 
& 0.860 (0.839) & 0.883 (0.715) & 0.810 (0.695) \\

& Qwen 3 32B & 0.827 (0.740) & 0.729 (0.506) & 0.799 (0.691) & \underline{0.886 (0.870)} & 0.886 (0.851) & \textbf{0.888 (0.867)}
& 0.871 (0.817) & 0.880 (0.684) & 0.819 (0.717) \\

\midrule

\multirow{5}{1.4cm}{{AMC-23}} 
& Claude 4.5 Sonnet & 0.967 (0.997) & 0.969 (0.987) & 0.956 (0.960) & \textbf{0.972 (1.000)} & \underline{0.972 (1.000)} & 0.971 (1.000)
& 0.955 (1.00) & 0.900 (0.604) & 0.853 (0.588) \\

& DeepSeek R1 & 0.947 (0.749) & 0.953 (0.543) & 0.914 (0.229) & \underline{0.985 (1.000)} & 0.985 (1.000) & 0.985 (1.000)
& \textbf{0.985 (1.00)} & 0.877 (0.412) & 0.880 (0.592) \\

& GPT-OSS 20B & 0.948 (0.737) & 0.929 (0.507) & 0.948 (0.603) & \textbf{0.980 (1.000)} & \underline{0.980 (1.000)} & 0.980 (1.000) 
& 0.980 (1.00) & 0.980 (0.583) & 0.884 (0.630) \\

& Amazon Nova Premier & 0.526 (0.811) & 0.558 (0.486) & 0.570 (0.757) & \textbf{0.715 (1.000)} & 0.710 (1.000) & \underline{0.714 (1.000)}
& 0.584 (1.00) & 0.419 (0.362) & 0.351 (0.410) \\

& Qwen 3 32B & 0.610 (0.755) & 0.516 (0.561) & 0.610 (0.727) & 0.818 (1.000) & \textbf{0.821 (1.000)} & \underline{0.820 (1.000)} 
& 0.810 (1.00) & 0.637 (0.389) & 0.513 (0.299) \\

\midrule

\multirow{5}{1.4cm}{{AIME-24}} 
& Claude 4.5 Sonnet & 0.557 (0.932) & 0.620 (0.955) & 0.555 (0.909) & 0.681 (1.000) & \textbf{0.699 (1.000)} & \underline{0.682 (1.000)} 
& 0.567 (0.972) & 0.273 (0.144) & 0.450 (0.640) \\

& DeepSeek R1 & 0.857 (0.843) & 0.813 (0.455) & 0.891 (0.909) & 0.925 (1.0) & 0.925 (1.0) & \underline{0.925 (1.0)} 
& \textbf{0.917 (1.00)} & 0.799 (0.298) & 0.788 (0.715) \\

& GPT-OSS 20B & 0.601 (0.742) & 0.757 (0.648) & 0.757 (0.786) & 0.893 (1.000) & 0.904 (1.000) & 0.897 (1.000) & \underline{0.905 (1.000)}  & \textbf{0.891 (0.319)} & 0.876 (0.681) \\

& Amazon Nova Premier & \underline{0.233 (—)} & 0.231 (—) & 0.194 (—) & 0.196 (—) & 0.205 (—) & 0.191 (—) & 0.174 (—) & — & \textbf{0.294 (—)} \\

& Qwen 3 32B & 0.306 (—) & 0.230 (—) & 0.287 (—) & \underline{0.423 (—)} & 0.420 (—) & \textbf{0.426 (—)} & 0.414 (—) & — & 0.247 (—) \\

\midrule

\multirow{5}{1.4cm}{{AIME-25}} 
& Claude 4.5 Sonnet
& 0.580 (0.988) & 0.599 (1.00) & 0.204 (0.081)
& \underline{0.645 (1.00)} & 0.643 (1.00) & \textbf{0.646 (1.00)}
& 0.645 (1.00) & 0.489 (0.175) & 0.524 (0.263) \\

& DeepSeek R1
& 0.360 (0.630) & 0.380 (0.466) & 0.523 (0.397)
& 0.735 (1.00) & \underline{0.748 (1.00)} & \textbf{0.754 (1.00)}
& 0.707 (1.00) & 0.564 (0.400) & 0.616 (0.296) \\

& GPT-OSS 20B
& 0.424 (0.546) & 0.407 (0.370) & 0.634 (0.417)
& 0.880 (1.00) & \textbf{0.886 (1.00)} & \underline{0.886 (1.00)}
& 0.825 (1.00) & 0.776 (0.370) & 0.768 (0.491) \\

& Amazon Nova Premier
& 0.083 (--) & 0.126 (--) & 0.148 (--)
& 0.188 (--) & 0.192 (--) & \underline{0.220 (--)}
& 0.178 (--) & 0.106 (--) & \textbf{0.257 (--)} \\

& Qwen 3 32B
& 0.086 (0.759) & 0.134 (0.931) & 0.163 (0.517)
& 0.260 (1.00) & \underline{0.290 (1.00)} & \textbf{0.313 (1.00)}
& 0.208 (1.00) & 0.192 (0.897) & 0.126 (0.931) \\

\midrule
\multirow{5}{1.4cm}{{MMLU-Pro}} 
& Claude 4.5 Sonnet & 0.921 (0.808) & 0.922 (0.797) & 0.911 (0.770) & \textbf{0.945 (0.910)} & \underline{0.945 (0.909)} & 0.944 (0.909) 
& 0.900 (0.884) & 0.944 (0.885) & 0.890 (0.602) \\

& DeepSeek R1 & 0.892 (0.671) & 0.862 (0.568) & 0.878 (0.643) & \textbf{0.941 (0.913)} & \underline{0.933 (0.901)} & 0.930 (0.907) 
& 0.882 (0.882) & 0.927 (0.796) & 0.870 (0.577) \\

& GPT-OSS 20B & 0.706 (0.565) & 0.725 (0.500) & 0.726 (0.541) & \textbf{0.869 (0.933)} & \underline{0.868 (0.931)} & 0.868 (0.931)  
& 0.830 (0.935) & 0.785 (0.631) & 0.774 (0.573) \\

& Amazon Nova Premier & 0.705 (0.463) & 0.699 (0.531) & 0.690 (0.458) & \textbf{0.849 (0.957)} & 0.823 (0.944) & \underline{0.844 (0.955)}
& 0.801 (0.945) & 0.820 (0.775) & 0.811 (0.713) \\

& Qwen 3 32B & 0.692 (0.640) & 0.638 (0.501) & 0.671 (0.592) & \textbf{0.820 (0.972)} & 0.808 (0.968) & \underline{0.817 (0.971)}
& 0.787 (0.966) & 0.728 (0.686) & 0.742 (0.724) \\

\midrule
\multirow{5}{1.4cm}{{HotpotQA}} 
& Claude 4.5 Sonnet 
& 0.663 (0.571) & 0.717 (0.619) & 0.667 (0.552) & 0.681 (0.608) & 0.730 (0.653) & 0.686 (0.598) & \underline{0.839 (0.656)} & \textbf{0.847 (0.700)} & 0.768 (0.576) \\
& DeepSeek R1  
& 0.754 (0.635) & 0.730 (0.552) & 0.760 (0.628) & 0.792 (0.727) & 0.780 (0.699) & 0.796 (0.715) & \underline{0.835 (0.658)} & 0.829 (0.708) & \textbf{0.852 (0.696)} \\
& GPT-OSS 20B  
& 0.759 (0.637) & 0.725 (0.554) & 0.748 (0.603) 
& \textbf{0.819 (0.756)} & \underline{0.815 (0.739)} & 0.811 (0.743) 
& 0.813 (0.721) & 0.772 (0.592) & 0.80 (0.707) \\
& Amazon Nova Premier 
& 0.825 (0.606) & 0.779 (0.526) & 0.800 (0.571) 
& \textbf{0.866 (0.763)} & 0.835 (0.734) & 0.862 (0.760)
& \underline{0.864 (0.740)} & 0.812 (0.647) & 0.830 (0.618) \\
& Qwen 3 32B 
& 0.681 (0.620) & 0.659 (0.561) & 0.680 (0.613) 
& 0.761 (0.753) & \textbf{0.794 (0.770)} & 0.782 (0.770) 
& \underline{0.787 (0.729)} & 0.702 (0.630) & 0.707 (0.612) \\

\midrule
\multirow{5}{1.4cm}{{TruthfulQA}} 
& Claude 4.5 Sonnet & 0.997 (0.944) & 0.996 (0.926) & 0.994 (0.926) & 0.997 (0.955) & \textbf{0.997 (0.964)} & 0.996 (0.949) & 0.994 (0.926) & \underline{0.997 (0.949)} & 0.979 (0.503) \\
& DeepSeek R1 & 0.931 (0.604) & 0.897 (0.538) & 0.940 (0.844) & \underline{0.954 (0.886)} & \textbf{0.957 (0.888)} & 0.926 (0.819) & 0.940 (0.844) & 0.949 (0.721) & 0.907 (0.530) \\
& GPT-OSS 20B & 0.872 (0.633) & 0.865 (0.675) & 0.916 (0.936) & \textbf{0.921 (0.953)} & \underline{0.918 (0.952)} & 0.909 (0.946) & 0.916 (0.936) & 0.864 (0.608) & 0.856 (0.525) \\
& Amazon Nova Premier & 0.920 (0.616) & 0.948 (0.741) & 0.946 (0.862) & \textbf{0.963 (0.931)} & 0.946 (0.877) & \underline{0.963 (0.919)} & 0.946 (0.862) & 0.953 (0.828) & 0.907 (0.596) \\
& Qwen 3 32B & 0.780 (0.553) & 0.752 (0.500) & 0.856 (0.936) & \textbf{0.865 (0.957)} & \underline{0.857 (0.946)} & 0.853 (0.950) & 0.856 (0.936) & 0.822 (0.694) & 0.812 (0.636) \\

\bottomrule
\end{tabular}%
}
\caption{\textbf{Selective prediction performance (Sel-AUC; AUROC in parentheses) -- Combined StructU+Self-ConsU.}
Sel-AUC measures area under the risk--coverage curve (higher is better).
Positive class = robust failure under sampling ($\tau{=}1.0$).
\textbf{StructU+Self-ConsU} reports the combined estimator (\textit{within}, \textit{across}, \textit{total}) using TrueSkill+PageRank as the preference backend.
Baselines: \textbf{Self-ConsU}, \textbf{VerbalizedU}, \textbf{SemanticU}.
Bold = best; underline = second-best per row.}
\label{tab:appendix_hybridstructu_detailed_results}
\end{table*}

\subsection{MMLU-Pro Domain Breakdown}

Table~\ref{tab:domain_results} reports performance across MMLU-Pro domains. In Physics and Math, StructU-within consistently outperforms other components, aligning with the intuition that these domains admit multiple valid derivations. In Engineering and Law, the hybrid variants show largest gains, suggesting structural rankings provide scaffolding that improves selective prediction when paired with self-consistency.

\begin{table*}[t]
\centering
\scriptsize
\setlength{\tabcolsep}{2.8pt}
\renewcommand{\arraystretch}{1.15}
\resizebox{\textwidth}{!}{%
\begin{tabular}{p{1.1cm}p{2.4cm}ccccccccc}
\toprule
\multirow{2}{1.1cm}{Domain} & \multirow{2}{2.4cm}{Model} &
\multicolumn{3}{c}{\cellcolor{blue!12}\textbf{StructU (Ours)}} &
\multicolumn{3}{c}{\cellcolor{red!12}\textbf{StructU+ConsU (Ours)}} &
\multicolumn{2}{c}{\textbf{Baselines}} \\
\cmidrule(lr){3-5}\cmidrule(lr){6-8}\cmidrule(lr){9-10}
& &
\cellcolor{blue!12}within & \cellcolor{blue!12}across & \cellcolor{blue!12}total &
\cellcolor{red!12}within & \cellcolor{red!12}across & \cellcolor{red!12}total &
Self-ConsU & SemanticU \\
\midrule

\multirow{5}{1.05cm}{Chemistry}
& Claude 4.5 Sonnet
& 0.928 (0.793) & 0.940 (0.782) & 0.842 (0.305)
& 0.945 (0.871) & \textbf{0.955 (0.890)} & \underline{0.950 (0.892)}
& 0.935 (0.867)  & 0.916 (0.583) \\
& DeepSeek R1
& 0.921 (0.729) & 0.856 (0.492) & 0.808 (0.328)
& \textbf{0.959 (0.931)} & \underline{0.939 (0.899)} & 0.915 (0.864)
& 0.937 (0.889)  & 0.860 (0.497) \\
& GPT-OSS 20B
& 0.812 (0.583) & 0.809 (0.493) & 0.840 (0.462)
& 0.901 (0.903) & \underline{0.903 (0.904)} & \textbf{0.929 (0.926)}
& 0.889 (0.911) & 0.836 (0.538) \\
& Amazon Nova Premier
& 0.739 (0.660) & 0.685 (0.451) & 0.701 (0.398)
& 0.872 (0.977) & \underline{0.877 (0.980)} & \textbf{0.886 (0.985)}
& 0.875 (0.982) & 0.821 (0.734) \\
& Qwen 3 32B
& 0.742 (0.680) & 0.716 (0.540) & 0.618 (0.344)
& \textbf{0.840 (0.980)} & 0.838 (0.978) & 0.834 (0.976)
& \underline{0.839 (0.977)}  & 0.764 (0.712) \\
\midrule

\multirow{5}{1.05cm}{Engineering}
& Claude 4.5 Sonnet
& 0.854 (0.813) & 0.853 (0.791) & 0.658 (0.194)
& \textbf{0.913 (0.946)} & \underline{0.906 (0.942)} & 0.893 (0.925)
& 0.899 (0.937)  & 0.803 (0.587) \\
& DeepSeek R1
& 0.829 (0.625) & 0.755 (0.453) & 0.791 (0.455)
& \textbf{0.950 (0.965)} & 0.931 (0.944) & 0.928 (0.943)
& \underline{0.943 (0.950)} & 0.849 (0.592) \\
& GPT-OSS 20B
& 0.464 (0.406) & 0.485 (0.363) & 0.675 (0.557)
& 0.834 (0.977) & \underline{0.839 (0.979)} & \textbf{0.847 (0.982)}
& 0.836 (0.978)  & 0.604 (0.516) \\
& Amazon Nova Premier
& 0.598 (0.596) & 0.529 (0.383) & 0.597 (0.447)
& 0.826 (0.981) & \underline{0.832 (0.989)} & \textbf{0.844 (0.990)}
& 0.813 (0.984)  & 0.750 (0.729) \\
& Qwen 3 32B
& 0.683 (0.652) & 0.642 (0.581) & 0.570 (0.376)
& \textbf{0.804 (0.977)} & \underline{0.802 (0.978)} & \textbf{0.804 (0.978)}
& 0.799 (0.978) & 0.688 (0.665) \\
\midrule

\multirow{5}{1.05cm}{Law}
& Claude 4.5 Sonnet
& 0.848 (0.803) & 0.818 (0.705) & 0.648 (0.268)
& \textbf{0.876 (0.879)} & \underline{0.853 (0.846)} & 0.803 (0.790)
& 0.832 (0.838)  & 0.723 (0.483) \\
& DeepSeek R1
& 0.717 (0.568) & 0.725 (0.552) & 0.695 (0.464)
& \underline{0.788 (0.826)} & \textbf{0.793 (0.822)} & 0.775 (0.809)
& 0.772 (0.822)  & 0.713 (0.529) \\
& GPT-OSS 20B
& 0.460 (0.619) & 0.437 (0.535) & 0.422 (0.443)
& \textbf{0.541 (0.914)} & 0.530 (0.908) & 0.530 (0.909)
& \underline{0.537 (0.908)}  & 0.450 (0.540) \\
& Amazon Nova Premier
& 0.471 (0.478) & 0.502 (0.516) & 0.542 (0.588)
& 0.581 (0.846) & \underline{0.620 (0.869)} & \textbf{0.643 (0.888)}
& 0.589 (0.861)  & 0.578 (0.599) \\
& Qwen 3 32B
& 0.454 (0.572) & 0.446 (0.578) & 0.431 (0.499)
& \textbf{0.571 (0.934)} & 0.553 (0.917) & 0.538 (0.910)
& \underline{0.554 (0.922)} & 0.463 (0.606) \\
\midrule

\multirow{5}{1.05cm}{Math}
& Claude 4.5 Sonnet
& 0.963 (0.839) & 0.956 (0.787) & 0.860 (0.243)
& \textbf{0.974 (0.927)} & \underline{0.973 (0.923)} & 0.959 (0.868)
& 0.950 (0.888) & 0.948 (0.650) \\
& DeepSeek R1
& 0.952 (0.711) & 0.945 (0.658) & 0.879 (0.353)
& \textbf{0.968 (0.926)} & \underline{0.965 (0.930)} & 0.950 (0.886)
& 0.945 (0.892) & 0.918 (0.556) \\
& GPT-OSS 20B
& 0.880 (0.567) & 0.891 (0.524) & 0.907 (0.479)
& 0.960 (0.940) & \underline{0.962 (0.956)} & \textbf{0.966 (0.962)}
& 0.961 (0.944) & 0.904 (0.548) \\
& Amazon Nova Premier
& 0.796 (0.668) & 0.779 (0.550) & 0.762 (0.441)
& \textbf{0.899 (0.961)} & \underline{0.896 (0.968)} & 0.894 (0.957)
& 0.878 (0.958)  & 0.849 (0.689) \\
& Qwen 3 32B
& 0.815 (0.653) & 0.812 (0.637) & 0.726 (0.442)
& \underline{0.915 (0.981)} & \textbf{0.921 (0.995)} & \underline{0.915 (0.988)}
& 0.894 (0.982)  & 0.796 (0.577) \\
\midrule

\multirow{5}{1.05cm}{Physics}
& Claude 4.5 Sonnet
& 0.955 (0.822) & 0.956 (0.825) & 0.832 (0.207)
& \underline{0.967 (0.910)} & \textbf{0.969 (0.922)} & 0.958 (0.891)
& 0.965 (0.893)  & 0.919 (0.577) \\
& DeepSeek R1
& 0.921 (0.649) & 0.918 (0.585) & 0.890 (0.470)
& 0.959 (0.923) & \underline{0.962 (0.925)} & 0.960 (0.925)
& \textbf{0.963 (0.907)} & 0.895 (0.542) \\
& GPT-OSS 20B
& 0.789 (0.530) & 0.800 (0.480) & 0.811 (0.420)
& 0.907 (0.933) & 0.921 (0.943) & \textbf{0.935 (0.958)}
& \underline{0.930 (0.943)}  & 0.813 (0.484) \\
& Amazon Nova Premier
& 0.787 (0.691) & 0.705 (0.454) & 0.721 (0.413)
& 0.887 (0.964) & \underline{0.893 (0.968)} & \textbf{0.901 (0.973)}
& 0.889 (0.965) & 0.833 (0.742) \\
& Qwen 3 32B
& 0.793 (0.750) & 0.739 (0.599) & 0.676 (0.405)
& 0.874 (0.979) & \underline{0.875 (0.979)} & \underline{0.875 (0.977)}
& \textbf{0.879 (0.976)} & 0.813 (0.758) \\
\bottomrule
\end{tabular}%
}
\caption{\textbf{Selective prediction (Sel-AUC; AUROC in parentheses).}
Higher is better; positive class = robust failure ($\tau{=}1.0$).
\textbf{StructU} reports structural uncertainty (\textit{within}, \textit{across}, \textit{total}), selecting the best variant among Bradley--Terry+PageRank and TrueSkill+PageRank.
\textbf{StructU+Self-ConsU} reports the combined estimators.
Baselines: \textbf{Self-ConsU}, \textbf{SemanticU}.
Bold = best; underline = second-best per row.}
\label{tab:domain_results}
\end{table*}

\subsection{Robustness to Correctness Thresholds}

Table~\ref{tab:thresholds_per_model} reports AUROC under three correctness thresholds $\tau\in\{1.0,0.8,0.6\}$. All uncertainty signals exhibit modest degradation as the threshold is relaxed, confirming that separability is not an artifact of strict labeling but remains stable under looser correctness definitions.

\begin{table*}[t]
\centering
\small
\setlength{\tabcolsep}{4pt}
\renewcommand{\arraystretch}{1.05}
\resizebox{\textwidth}{!}{%
\begin{tabular}{llccc ccc ccc}
\toprule
\multirow{2}{*}{\textbf{Dataset}} & \multirow{2}{*}{\textbf{Model}} &
\multicolumn{3}{c}{\textbf{Self-ConsU}} &
\multicolumn{3}{c}{\textbf{StructU\_within (Bradley--Terry+PageRank)}} &
\multicolumn{3}{c}{\textbf{StructU+Self-ConsU\_within}} \\
\cmidrule(lr){3-5}\cmidrule(lr){6-8}\cmidrule(lr){9-11}
& & $\tau{=}1.0$ & $\tau{=}0.8$ & $\tau{=}0.6$
  & $\tau{=}1.0$ & $\tau{=}0.8$ & $\tau{=}0.6$
  & $\tau{=}1.0$ & $\tau{=}0.8$ & $\tau{=}0.6$ \\
\midrule

\multirow{5}{*}{\textbf{Math-Synth}}
& Claude 4.5 Sonnet   & 0.978 & 0.978 & 0.846 & 0.986 & 0.905 & 0.773 & 0.992 & 0.977 & 0.846 \\
& DeepSeek R1         & 0.889 & 0.889 & 0.798 & 0.902 & 0.853 & 0.770 & 0.931 & 0.916 & 0.798 \\
& GPT-OSS 20B         & 0.918 & 0.918 & 0.792 & 0.775 & 0.779 & 0.710 & 0.955 & 0.920 & 0.792 \\
& Amazon Nova Premier & 0.986 & 0.986 & 0.911 & 0.938 & 0.912 & 0.846 & 0.997 & 0.987 & 0.911 \\
& Qwen 3 32B          & 0.990 & 0.990 & 0.862 & 0.850 & 0.766 & 0.710 & 0.998 & 0.991 & 0.862 \\
\midrule

\multirow{5}{*}{\textbf{MATH-500}}
& Claude 4.5 Sonnet   & 0.818 & 0.811 & 0.775 & 0.814 & 0.734 & 0.681 & 0.842 & 0.807 & 0.760 \\
& DeepSeek R1         & 0.765 & 0.758 & 0.726 & 0.631 & 0.596 & 0.554 & 0.768 & 0.757 & 0.717 \\
& GPT-OSS 20B         & 0.705 & 0.679 & 0.618 & 0.577 & 0.534 & 0.482 & 0.706 & 0.686 & 0.618 \\
& Amazon Nova Premier & 0.876 & 0.852 & 0.824 & 0.702 & 0.751 & 0.714 & 0.874 & 0.858 & 0.818 \\
& Qwen 3 32B          & 0.854 & 0.872 & 0.842 & 0.767 & 0.795 & 0.742 & 0.886 & 0.894 & 0.814 \\
\midrule

\multirow{5}{*}{\textbf{MMLU-Pro}}
& Claude 4.5 Sonnet   & 0.884 & 0.833 & 0.783 & 0.834 & 0.768 & 0.728 & 0.912 & 0.865 & 0.819 \\
& DeepSeek R1         & 0.882 & 0.813 & 0.757 & 0.559 & 0.502 & 0.473 & 0.888 & 0.818 & 0.760 \\
& GPT-OSS 20B         & 0.935 & 0.870 & 0.812 & 0.456 & 0.394 & 0.380 & 0.918 & 0.850 & 0.785 \\
& Amazon Nova Premier & 0.945 & 0.846 & 0.735 & 0.610 & 0.506 & 0.464 & 0.933 & 0.838 & 0.723 \\
& Qwen 3 32B          & 0.966 & 0.904 & 0.799 & 0.673 & 0.660 & 0.586 & 0.970 & 0.911 & 0.796 \\
\bottomrule
\end{tabular}%
}
\caption{\emph{Robustness to correctness thresholds (per-model AUROC).}
AUROC for detecting \emph{robust failures under sampling} at thresholds $\tau\in\{1.0,0.8,0.6\}$ with $N{=}5$.
A question is labeled incorrect if $\hat{p}_{\mathrm{corr}}<\tau$.
We report: Self-ConsU, StructU\_within (within-trial from Bradley--Terry+PageRank), and the combined StructU+Self-ConsU\_within.
Higher is better.}
\label{tab:thresholds_per_model}
\end{table*}

\section{Additional Analysis of Structural Uncertainty Signals}
\label{app:qualitative}

\subsection{Structural Collapse Across Additional Models for HotpotQA}

\begin{figure*}[ht]
    \centering
    \includegraphics[width=\textwidth]{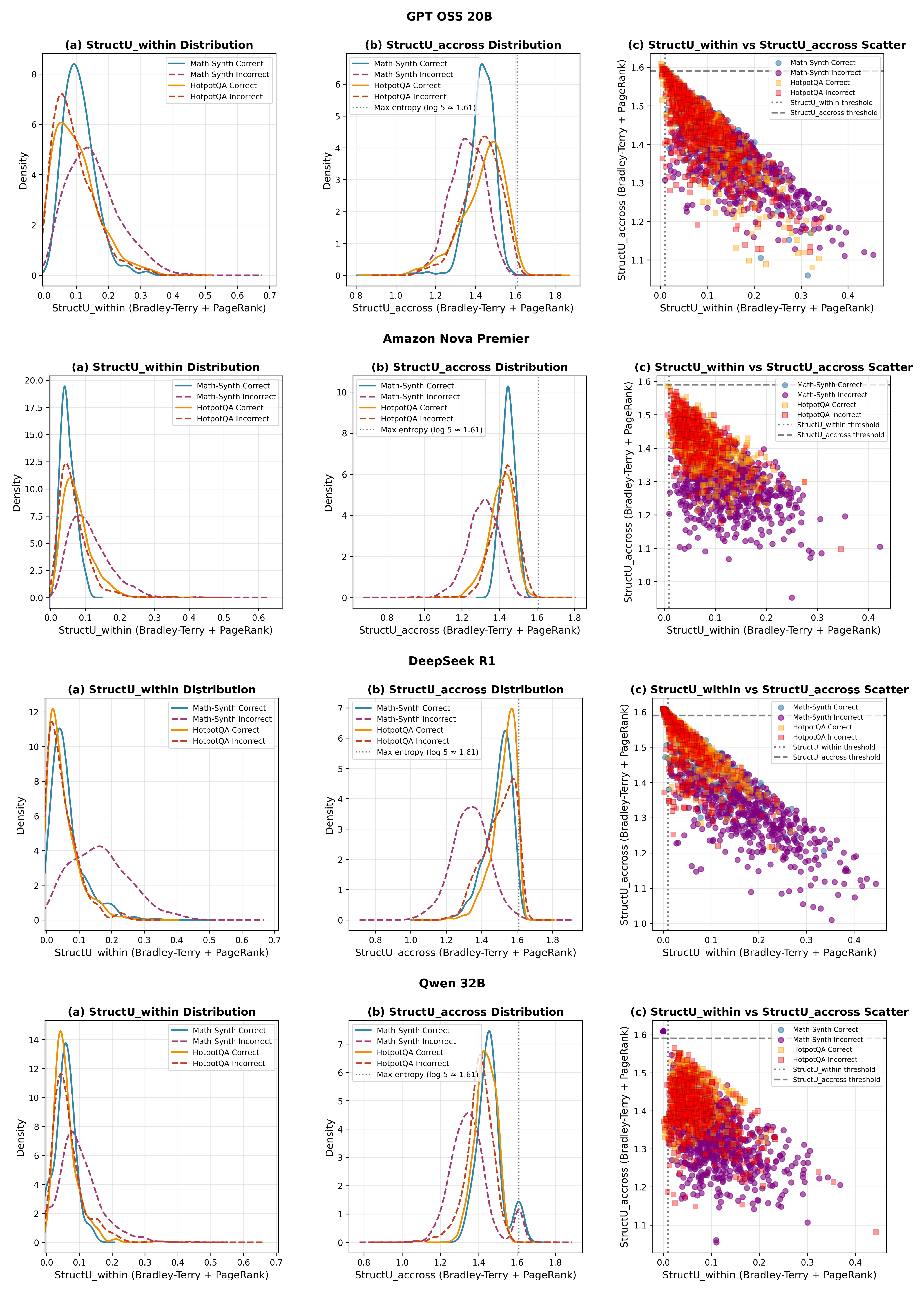}
    \caption{\textbf{Structural collapse on factual retrieval across four additional models.} Rows correspond to Amazon Nova Premier, DeepSeek R1, GPT-OSS 20B, and Qwen 3 32B. Each row shows the across-trial uncertainty distribution (left), within-trial uncertainty distribution (center), and joint across-trial--within-trial scatter (right) on Math-Synth and HotpotQA, conditioned on correctness (Bradley--Terry + PageRank). The ``HotpotQA signature''---near-zero across-trial uncertainty with near-maximum within-trial uncertainty ($\log 5 \approx 1.61$, dotted line)---is reproduced across all models, confirming that the structural collapse is a task-level phenomenon independent of model capability.}
    \label{fig:structural_collapse_additional}
\end{figure*}

Figure~\ref{fig:structural_collapse_additional} extends the structural collapse analysis to Amazon Nova Premier, DeepSeek R1, GPT-OSS 20B, and Qwen 3 32B. The ``HotpotQA signature''---near-zero across-trial uncertainty with near-maximum within-trial uncertainty ($\log 5 \approx 1.61$)---reproduces identically across all models, confirming that factual retrieval elicits prompt-invariant reasoning chains rendering preference graphs uninformative regardless of model capability. In contrast, Math-Synth retains clear across-trial separation between correct and incorrect questions for all models, though the dynamic range narrows for weaker models (Amazon Nova Premier, Qwen 3 32B).

This universality aligns with Table~\ref{tab:main_results}: on HotpotQA, StructU+Self-ConsU fails to improve over Self-ConsU alone for DeepSeek R1 (0.796 vs 0.835 Sel-AUC) and Claude 4.5 (0.742 vs 0.839), confirming degenerate preference graphs introduce noise rather than signal.

\subsection{Qualitative Assessment of Structural Collapse}
To provide mechanistic evidence for the quantitative findings in Section~\ref{sec:results}, we analyze four examples from Claude 4.5 Sonnet with $\text{Self-ConsU} = 0$ (unanimous agreement), isolating structural signals where dispersion-based methods are uninformative. Table~\ref{tab:qualitative_summary} summarizes key quantities.

\begin{table*}[t]
\centering
\small
\begin{tabular}{llcccccl}
\toprule
\textbf{Dataset} & \textbf{Correct?} & \textbf{Self-ConsU} & $\text{StructU}_{\text{across}}$ & $\text{StructU}_{\text{within}}$ & $\pi_{\max}/\pi_{\min}$ & $\text{CV}_{\max}$ & \textbf{Diagnosis} \\
\midrule
Math-Synth & \xmark & 0.0 & \textbf{0.035} & 1.549 & 1.73 & 0.497 & Signal fires \\
Math-Synth & \cmark & 0.0 & 0.001 & 1.607 & 1.09 & 0.047 & Appropriately quiet \\
HotpotQA   & \xmark & 0.0 & $<$0.001 & 1.609 & 1.06 & 0.013 & Collapsed \\
HotpotQA   & \cmark & 0.0 & $<$0.001 & 1.608 & 1.08 & 0.015 & Collapsed \\
\bottomrule
\end{tabular}
\caption{Summary of qualitative examples. All four satisfy $\text{Self-ConsU} = 0$. $\text{StructU}_{\text{across}}$: across-trial uncertainty; $\text{StructU}_{\text{within}}$: within-trial uncertainty. $\pi_{\max}/\pi_{\min}$: ratio of largest to smallest mean PageRank score. $\text{CV}_{\max}$: maximum coefficient of variation of PageRank across spanning tree trials.}
\label{tab:qualitative_summary}
\end{table*}

\subsection{Math-Synth: Structural Diversity Enables Discrimination}
\label{app:qual_math}

\subsubsection{Incorrect Example ($\text{Self-ConsU} = 0$, $\text{StructU}_{\text{across}} = 0.035$)}

\paragraph{Task.} A 6-digit synthetic arithmetic problem involving nested negations and multiplication:

\begin{equation*}
\resizebox{0.75\columnwidth}{!}{$
\underbrace{-(-(-(-(-(-}_{6\text{ outer negations}}
\underbrace{-(-500)}_{=500} \times 200
\;)\;)\;)\;)\;)\;)
\;\;
\underbrace{- -(-(-1))}_{=1}
$}
\end{equation*}

\noindent The left sub-expression evaluates as $-(-500)\times 200 = 100{,}000$, then wrapped in \textbf{6 outer negations} (even count $\to$ positive), yielding $+100{,}000$. Adding the right part: $100{,}000 + 1 = 100{,}001$. All five responses unanimously answer $-99{,}999$; the ground truth is $\mathbf{100{,}001}$.

\paragraph{Shared error.} All responses miscount outer negations as \textbf{5} (odd $\rightarrow$ negative) instead of \textbf{6} (even $\rightarrow$ positive), flipping the sign from $+100{,}000$ to $-100{,}000$ via different error paths (Table~\ref{tab:math_incorrect_traces}).

\begin{table*}[t]
\centering
\small
\renewcommand{\arraystretch}{1.3}
\begin{tabular}{p{0.5cm}p{1.7cm}p{9.2cm}c}
\toprule
\textbf{ID} & \textbf{Strategy} & \textbf{Key Reasoning Steps} & $\bar{\pi}$ \\
\midrule
R1 & Step-by-step w/ self-check &
$-(-500) \times 200 = 100{,}000$ $\;\rightarrow\;$ counts 7 total negation signs $\;\rightarrow\;$ \textcolor{red}{counts 5 remaining after multiply (correct: 6)} $\;\rightarrow\;$ \textcolor{red}{odd $\rightarrow$ $-100{,}000$} \;[correct: \textcolor{ForestGreen}{even $\rightarrow$ $+100{,}000$}] $\;\rightarrow\;$ self-check re-derives each step but \emph{repeats same miscount} $\;\rightarrow\;$ $-100{,}000 + 1 = -99{,}999$
& 0.238 \\
\midrule
R2 & Think-aloud &
Metacognitive narration: ``I need to count remaining negations after multiplication'' $\;\rightarrow\;$ \textcolor{red}{identifies 5 remaining (correct: 6)} $\;\rightarrow\;$ \textcolor{red}{$-100{,}000$} \;[correct: \textcolor{ForestGreen}{$+100{,}000$}] $\;\rightarrow\;$ $-100{,}000 + 1 = -99{,}999$
& 0.215 \\
\midrule
R3 & Socratic &
Reframes as parity problem: \textcolor{red}{``7 negations total (odd) applied to $-500 \times 200 = -100{,}000$ gives $-100{,}000$''} --- conflates inner negation with outer count \;[correct: \textcolor{ForestGreen}{6 outer negations (even) applied to $+100{,}000$}] $\;\rightarrow\;$ $-100{,}000 + 1 = -99{,}999$
& 0.224 \\
\midrule
R4 & Decomposition &
Sub-problem~1: left expression $\;\rightarrow\;$ Sub-problem~2: right expression $\;\rightarrow\;$ \textcolor{red}{$-100{,}000$} \;[correct: \textcolor{ForestGreen}{$+100{,}000$}] $+ 1 = -99{,}999$. Most modular structure; \emph{no negation count shown, no verification step}.
& 0.138 \\
\midrule
R5 & Analogical reasoning &
Relates to simpler cases ($-(-5)=5$, $-(-(-3))=-3$) $\;\rightarrow\;$ \textcolor{red}{parses $-500 \times 200 = -100{,}000$ then applies 6 negations to $-100{,}000$} \;[correct: \textcolor{ForestGreen}{$-(-500) \times 200 = +100{,}000$ then 6 negations}] $\;\rightarrow\;$ initial miscount, self-corrects counting but retains inner parsing error $\;\rightarrow\;$ $-100{,}000 + 1 = -99{,}999$
& 0.185 \\
\bottomrule
\end{tabular}
\caption{Reasoning traces for the Math-Synth incorrect example. The correct evaluation requires \textbf{6 outer negations} (even $\rightarrow$ \textcolor{ForestGreen}{$+100{,}000$}), but all responses miscount \textbf{5} (odd $\rightarrow$ \textcolor{red}{$-100{,}000$}). Each response reaches the same wrong answer via a structurally different error path. Ground truth: $100{,}001$; unanimous model answer: $-99{,}999$.}
\label{tab:math_incorrect_traces}
\end{table*}

\paragraph{PageRank dynamics.} The structural diversity across responses produces a skewed PageRank distribution ($\pi_{\max}/\pi_{\min} = 1.73$) with substantial instability across spanning tree trials. Table~\ref{tab:math_incorrect_pagerank} reports the per-trial PageRank vectors.

\begin{table}[t]
\centering
\small
\begin{tabular}{lccccc}
\toprule
& R1 & R2 & R3 & R4 & R5 \\
\midrule
Trial 0 & 0.263 & 0.252 & 0.243 & \textbf{0.055} & 0.186 \\
Trial 1 & 0.225 & 0.190 & 0.212 & 0.212 & 0.161 \\
Trial 2 & 0.205 & 0.219 & 0.189 & 0.164 & 0.224 \\
Trial 3 & 0.265 & 0.233 & 0.255 & \textbf{0.057} & 0.190 \\
Trial 4 & 0.233 & 0.180 & 0.222 & 0.202 & 0.163 \\
\midrule
Mean    & 0.238 & 0.215 & 0.224 & 0.138 & 0.185 \\
CV      & 0.097 & 0.124 & 0.103 & \textbf{0.497} & 0.124 \\
\bottomrule
\end{tabular}
\caption{Per-trial PageRank distributions for the Math-Synth incorrect example. Response~4 (decomposition, no verification) exhibits extreme instability (CV $= 0.497$), fluctuating between $0.055$ and $0.212$ across trials.}
\label{tab:math_incorrect_pagerank}
\end{table}

\paragraph{Analysis.} Responses differ substantively in verification depth (R1: explicit self-check; R4: none), metacognitive structure (R2: narrated strategy; R5: mid-derivation correction), and error pathway (R1--R4: miscount negations; R5: misparse grouping)---genuine derivation quality differences, not surface reformulations.

Table~\ref{tab:math_incorrect_prefs} shows these differences produce unstable preferences. Across 20 judgments, 80\% of confidence scores are $\leq$65, indicating near-indifference. The R1 vs R2 pair reveals instability: the judge cites "clarity" to prefer R1 in trials 3--4 but R2 in trials 1 and 5. R4's PageRank fluctuates between 0.055 and 0.212 (CV=0.497) because ranking depends on which spanning tree edges are sampled.

\begin{table}[t]
\centering
\footnotesize
\setlength{\tabcolsep}{2.5pt}
\renewcommand{\arraystretch}{1.2}
\begin{tabular}{@{}llc>{\raggedright\arraybackslash}p{3.2cm}@{}}
\toprule
\textbf{Pair} & \textbf{Winner Trials} & \textbf{Conf.} & \textbf{Instability} \\
\midrule
R1 vs R2 & Flips: 3,4$\succ$1,5 & 55--62 & "Clarity" cited both ways \\
R4 vs R5 & R5: 3,4 / R4: 1 & 65--85 & R4 rejected (85) vs R5, marginal (55) vs R1 \\
R1 vs R5 & R1: 2,3,4 stable & 72 & R1 self-check valued over R5 correction \\
\bottomrule
\end{tabular}
\caption{Judge preference instability (Math-Synth incorrect). Confidence: 50=no preference, 100=certainty.}
\label{tab:math_incorrect_prefs}
\end{table}

This instability maps onto the two uncertainty components. \textbf{Across-trial} ($\text{StructU}_{\text{across}} = 0.035$): competing quality criteria (verification depth, metacognitive clarity, modularity) produce different winners depending on sampled edges, shifting PageRank vectors between trials. \textbf{Within-trial} ($\text{StructU}_{\text{within}} = 1.549$): low confidence ($\leq$65) prevents domination, distributing PageRank mass across multiple candidates per trial. Together, elevated $\text{StructU}_{\text{across}}$ correctly flags unreliability despite $\text{Self-ConsU} = 0$, while high $\text{StructU}_{\text{within}}$ reflects multiple distinct---though uniformly flawed---reasoning strategies.

\subsubsection{Correct Example ($\text{Self-ConsU} = 0$, $\text{StructU}_{\text{across}} \approx 0.001$)}

\paragraph{Task.} A 14-digit synthetic arithmetic problem involving nested negations and multiplication:
\begin{equation*}
\resizebox{0.75\columnwidth}{!}{$
\underbrace{-(-(-(-(-(-}_{6\text{ outer negations}}
(\underbrace{-9\,999\,999\,999\,999}_{13\text{ nines}}
\times -1) + 1
\;)\;)\;)\;)\;)\;)
$}
\end{equation*}

\noindent The inner product evaluates to $({-9\,999\,999\,999\,999})\times({-1}) = 9\,999\,999\,999\,999$. Adding $1$ gives $10\,000\,000\,000\,000$. Six outer negations (even count $\to$ sign unchanged) yield $+10\,000\,000\,000\,000$. All five responses unanimously answer $10\,000\,000\,000\,000$, matching the ground truth.

\paragraph{All responses correct.} Every response reaches the correct answer through sound reasoning. The five prompt templates produce identical arithmetic but differ in presentation: R1 adds self-check with (\textit{verified}) annotations; R2 narrates thinking with even/odd shortcut; R3 frames as Socratic Q\&A; R4 decomposes into sub-problems; R5 draws analogy to $-(-(-3))$. Because computation is straightforward and all strategies succeed, variation is purely stylistic (Table~\ref{tab:math_correct_traces}).

\begin{table*}[t]
\centering
\small
\renewcommand{\arraystretch}{1.3}
\begin{tabular}{p{0.5cm}p{1.3cm}p{8cm}c}
\toprule
\textbf{ID} & \textbf{Strategy} & \textbf{Key Reasoning Steps} & $\bar{\pi}$ \\
\midrule
R1 & Step-by-step w/ self-check &
  $(-9\!\cdots\!9)\times(-1) = 9\!\cdots\!9$ $\;\rightarrow\;$ adds 1 to get $10^{13}$ $\;\rightarrow\;$ \textcolor{ForestGreen}{counts 6 outer negations (even $\rightarrow$ positive)} $\;\rightarrow\;$ applies each negation step-by-step (Steps~3--8) $\;\rightarrow\;$ \emph{self-check} recounts negation signs from original expression, re-verifies each step with \checkmark{} marks $\;\rightarrow\;$ \textcolor{ForestGreen}{$10{,}000{,}000{,}000{,}000$}
& 0.220 \\
\midrule
R2 & Think-aloud &
  Metacognitive narration: ``I need to work from the innermost parentheses outward'' $\;\rightarrow\;$ same arithmetic $\;\rightarrow\;$ \textcolor{ForestGreen}{counts 6 negative signs} $\;\rightarrow\;$ applies each negation $\;\rightarrow\;$ notes ``6 negations = even = positive result'' $\;\rightarrow\;$ \textcolor{ForestGreen}{$10{,}000{,}000{,}000{,}000$}
& 0.215 \\
\midrule
R3 & Socratic &
  ``What is the problem asking?'' \ldots\ ``What method should I use?'' $\;\rightarrow\;$ \textcolor{ForestGreen}{identifies 6 consecutive negation operations} $\;\rightarrow\;$ applies each negation (Steps~3--8) $\;\rightarrow\;$ ``Does my answer make sense? Yes.\ 6 negations (even) $\rightarrow$ positive'' $\;\rightarrow\;$ \textcolor{ForestGreen}{$10{,}000{,}000{,}000{,}000$}
& 0.195 \\
\midrule
R4 & Decomposition &
  Sub-problem~1: inner product $\;\rightarrow\;$ Sub-problem~2: add~1 $\;\rightarrow\;$ Sub-problem~3: apply 6 negations $\;\rightarrow\;$ \textcolor{ForestGreen}{``even number of negations (6), the result is positive''} $\;\rightarrow\;$ \textcolor{ForestGreen}{$10{,}000{,}000{,}000{,}000$}. Most modular structure; concise summary paragraph.
& 0.210 \\
\midrule
R5 & Analogical reasoning &
  Relates to simpler cases: ``similar to evaluating $-(-(-3))$'' $\;\rightarrow\;$ states $-(-x)=x$ principle $\;\rightarrow\;$ same step-by-step negations $\;\rightarrow\;$ \textcolor{ForestGreen}{``6 is even, result is positive''} $\;\rightarrow\;$ \textcolor{ForestGreen}{$10{,}000{,}000{,}000{,}000$}
& 0.160 \\
\bottomrule
\end{tabular}
\caption{Reasoning traces for the Math-Synth correct example. The correct evaluation requires \textbf{6 outer negations} (even $\rightarrow$ \textcolor{ForestGreen}{$+10^{13}$}), and all responses count correctly. Each response reaches the same right answer via a stylistically different but arithmetically equivalent path. Ground truth: $10{,}000{,}000{,}000{,}000$; unanimous model answer: $10{,}000{,}000{,}000{,}000$.}
\label{tab:math_correct_traces}
\end{table*}

\paragraph{PageRank dynamics.} The purely stylistic variation across responses produces a compressed PageRank distribution ($\pi_{\max}/\pi_{\min} = 1.38$) with minimal instability across spanning tree trials. Table~\ref{tab:math_correct_pagerank} reports the per-trial PageRank vectors.

\begin{table}[t]
\centering
\small
\begin{tabular}{lccccc}
\toprule
       & R1    & R2    & R3    & R4    & R5    \\
\midrule
Trial 0 & 0.225 & 0.240 & 0.155 & 0.235 & 0.145 \\
Trial 1 & 0.230 & 0.220 & 0.185 & 0.215 & 0.150 \\
Trial 2 & 0.215 & 0.210 & 0.205 & 0.205 & 0.165 \\
Trial 3 & 0.225 & 0.215 & 0.180 & 0.210 & 0.170 \\
Trial 4 & 0.220 & 0.210 & 0.200 & 0.210 & 0.160 \\
\midrule
Mean    & 0.220 & 0.215 & 0.195 & 0.210 & 0.160 \\
CV      & 0.024 & 0.054 & 0.089 & 0.054 & 0.060 \\
\bottomrule
\end{tabular}
\caption{Per-trial PageRank distributions for the Math-Synth correct example. All responses exhibit low coefficient of variation; compare R4's CV\,$=$\,0.054 here with CV\,$=$\,0.497 in the incorrect example (Table~\ref{tab:math_incorrect_pagerank}).}
\label{tab:math_correct_pagerank}
\end{table}

\paragraph{Analysis.} Responses differ only in expository format: verification depth (R1: explicit self-check; R4: summary paragraph), pedagogical framing (R3: Socratic dialogue; R5: simpler analogues), and narrative style (R2: thinking-aloud). Unlike the incorrect example, these presentational differences lack substantive reasoning quality differences—every response counts negations correctly and arrives at the correct answer.

Table~\ref{tab:math_correct_prefs} shows these stylistic-only differences produce \emph{stable} preferences. Across 20 judgments, \textbf{zero reversals} occur. The judge applies consistent tie-breaking: explicit self-verification valued over conciseness (R1$\succ$R2), directness over pedagogical framing (R4$\succ$R3), both over analogical scaffolding (R5 ranked last). Confidence clusters at 52--62, reflecting genuine discrimination difficulty that is \emph{stable} rather than \emph{variable}.

\begin{table}[t]
\centering
\footnotesize
\setlength{\tabcolsep}{2.5pt}
\renewcommand{\arraystretch}{1.2}
\begin{tabular}{@{}llc>{\raggedright\arraybackslash}p{3.5cm}@{}}
\toprule
\textbf{Pair} & \textbf{Winner} & \textbf{Conf.} & \textbf{Pattern} \\
\midrule
R1 vs R5 & R1: 1,3,4,5 & 55 & Stable; "self-check adds rigor" \\
R1 vs R2 & R1: 1,2,4 & 55 & Stable; "verification" over "narration" \\
R4 vs R5 & R4: 3,4,5 & 52 & Stable; "decomposition more focused" \\
\bottomrule
\end{tabular}
\caption{Judge preference stability (Math-Synth correct). Zero reversals. Confidence: 50=no preference, 100=certainty.}
\label{tab:math_correct_prefs}
\end{table}

This stability maps onto the uncertainty components. \textbf{Across-trial} ($\text{StructU}_{\text{across}} \approx 0.001$) is near zero because stylistic preferences—however weakly held—are reproducible: the same criterion applied in the same direction every trial (Table~\ref{tab:math_correct_prefs}), so PageRank vectors barely shift (Table~\ref{tab:math_correct_pagerank}, all CVs$\leq$0.089). \textbf{Within-trial} ($\text{StructU}_{\text{within}}$) remains moderate because low confidence (52--62) prevents single-response domination. The near-zero $\text{StructU}_{\text{across}}$ correctly identifies reliability despite $\text{Self-ConsU} = 0$. The $\approx 30\times$ difference from the incorrect example ($0.001$ vs $0.035$) demonstrates the core claim: when reliably right, preferences are stable; when reliably wrong, preferences destabilize, even with identical surface agreement.

\subsection{HotpotQA: Preference Graph Collapse}
\label{app:qual_hotpot}

\subsubsection{Incorrect Example ($\text{Self-ConsU} = 0$, $\text{StructU}_{\text{across}} < 0.001$)}

\paragraph{Task.} A multi-hop question from HotpotQA: \emph{``Text Me Merry Christmas'' is a song performed by Kristen Bell and a group that originated at what university?} Expected reasoning: (i) identify the group as \textbf{Straight No Chaser} from retrieved passage (Document 10), then (ii) locate the group's university of origin. \textbf{Ground truth:} Indiana University. All five responses unanimously state the provided documents lack this information.

\paragraph{Shared failure.} All responses correctly identify Straight No Chaser from Document 10 (hop 1) but fail hop 2 due to missing context describing the group's origins. Every response correctly reports the answer cannot be determined—\textbf{unanimous incorrect agreement} ($\text{Self-ConsU} = 0$) driven by shared retrieval gap, not reasoning error. Table~\ref{tab:hotpot_incorrect_traces} shows identical retrieval chains despite different prompts, with variation limited to surface reformulation.

\begin{table*}[t]
\centering
\small
\renewcommand{\arraystretch}{1.3}
\begin{tabular}{p{0.5cm}p{1.7cm}p{9.2cm}c}
\toprule
\textbf{ID} & \textbf{Strategy} & \textbf{Key Reasoning Steps} & $\bar{\pi}$ \\
\midrule
R1 & Step-by-step w/ self-check &
  Identifies Doc~10 $\;\rightarrow\;$ extracts ``Straight No Chaser and Kristen Bell'' $\;\rightarrow\;$ searches all documents for university origin $\;\rightarrow\;$ \textcolor{red}{information not found} \;[correct: \textcolor{ForestGreen}{Indiana University}] $\;\rightarrow\;$ \emph{self-check} re-verifies Doc~10 is the only relevant source, confirms gap $\;\rightarrow\;$ abstains
& 0.202 \\
\midrule
R2 & Think-aloud &
  ``Let me read through the question carefully'' $\;\rightarrow\;$ identifies Doc~10 $\;\rightarrow\;$ scans Docs~1--9, lists each with one-line summary (``not relevant'') $\;\rightarrow\;$ \textcolor{red}{information not found} \;[correct: \textcolor{ForestGreen}{Indiana University}] $\;\rightarrow\;$ abstains
& 0.207 \\
\midrule
R3 & Socratic &
  ``What is the question asking?'' $\;\rightarrow\;$ identifies Doc~10 $\;\rightarrow\;$ ``none of the provided documents contain information about which university'' $\;\rightarrow\;$ \textcolor{red}{information not found} \;[correct: \textcolor{ForestGreen}{Indiana University}] $\;\rightarrow\;$ ``Does my answer make sense? \ldots\ the context only confirms they performed the song'' $\;\rightarrow\;$ abstains
& 0.197 \\
\midrule
R4 & Decomposition &
  Sub-question~1: identify group $\;\rightarrow\;$ Sub-question~2: find university $\;\rightarrow\;$ checks Docs~1--10 $\;\rightarrow\;$ \textcolor{red}{information not found} \;[correct: \textcolor{ForestGreen}{Indiana University}] $\;\rightarrow\;$ ``cannot answer based solely on provided documents'' $\;\rightarrow\;$ abstains
& 0.196 \\
\midrule
R5 & Analogical reasoning &
  Frames as two-step retrieval pattern: ``Entity~X associated with Entity~Y, find attribute of~Y'' $\;\rightarrow\;$ identifies Doc~10 $\;\rightarrow\;$ searches all documents $\;\rightarrow\;$ \textcolor{red}{information not found} \;[correct: \textcolor{ForestGreen}{Indiana University}] $\;\rightarrow\;$ abstains
& 0.198 \\
\bottomrule
\end{tabular}
\caption{Reasoning traces for the HotpotQA incorrect example. All five responses correctly identify \textbf{Straight No Chaser} as the group (hop~1) but fail hop~2 due to missing context. Each response reaches the same abstention via a stylistically different but substantively identical path. Ground truth: Indiana University; unanimous model answer: \emph{cannot be determined}.}
\label{tab:hotpot_incorrect_traces}
\end{table*}

\paragraph{PageRank dynamics.} Because responses are \emph{substantively identical}—each performs the same successful first hop and failed second hop—the judge has even less basis for discrimination than in Math-Synth correct. Table~\ref{tab:hotpot_incorrect_pagerank} confirms collapse: all trials produce near-identical, near-uniform distributions with maximum deviation from $1/N = 0.200$ of just $0.008$.

\begin{table}[t]
\centering
\small
\begin{tabular}{lccccc}
\toprule
       & R1    & R2    & R3    & R4    & R5    \\
\midrule
Trial 0 & 0.202 & 0.208 & 0.196 & 0.196 & 0.196 \\
Trial 1 & 0.201 & 0.207 & 0.201 & 0.195 & 0.195 \\
Trial 2 & 0.201 & 0.207 & 0.195 & 0.195 & 0.201 \\
Trial 3 & 0.202 & 0.208 & 0.196 & 0.196 & 0.196 \\
Trial 4 & 0.201 & 0.207 & 0.195 & 0.195 & 0.201 \\
\midrule
Mean    & 0.202 & 0.207 & 0.197 & 0.196 & 0.198 \\
CV      & 0.003 & 0.003 & 0.011 & 0.003 & 0.013 \\
\bottomrule
\end{tabular}
\caption{PageRank distributions across five responses for HotpotQA incorrect trials. Near-uniform distributions (max deviation from $1/N = 0.200$ of just $0.008$) confirm full rank collapse.}
\label{tab:hotpot_incorrect_pagerank}
\end{table}

\paragraph{Analysis.} Unlike Math-Synth incorrect---where responses reached the same wrong \emph{answer} via structurally different \emph{error paths}---here responses share both answer (abstention) and reasoning outcome (successful hop 1, failed hop 2). Variation is purely expository: R1 adds self-check, R2 lists documents, R3 uses Socratic Q\&A, R4 decomposes, and R5 frames as retrieval pattern.

Table~\ref{tab:hotpot_incorrect_prefs} shows the judge finds essentially nothing to discriminate. Across ${\sim}35$ judgments, \textbf{one reversal} occurs (R3 vs R4, iteration 4, conf=52). Confidence: 80\% at 52, R4 vs R5 at literal 50 (coin-flip) all appearances. Only outlier: R2 vs R5 at 62 citing R2's "explicit document-by-document listing"—the sole substantive distinction.

\begin{table}[t]
\centering
\footnotesize
\setlength{\tabcolsep}{2pt}
\renewcommand{\arraystretch}{1.2}
\begin{tabular}{@{}llc>{\raggedright\arraybackslash}p{3.5cm}@{}}
\toprule
\textbf{Pair} & \textbf{Winner} & \textbf{Conf.} & \textbf{Pattern} \\
\midrule
R2 vs R1 & R2: all trials & 52 & Stable; doc-by-doc listing preferred \\
R4 vs R5 & Tie: all trials & 50 & Literal coin-flip; "essentially a tie" \\
R3 vs R4 & R4: 2,5 / R3: 4 & 52--55 & \textbf{Only reversal} \\
\bottomrule
\end{tabular}
\caption{Judge preferences (HotpotQA incorrect). One reversal across ${\sim}35$ comparisons. Confidence: 50=no preference.}
\label{tab:hotpot_incorrect_prefs}
\end{table}

This near-total indifference maps onto uncertainty components. The collapse mechanism: factual retrieval over a fixed document set is deterministic. The model scans keywords, identifies documents, locates answer or does not. Different prompts cannot induce different retrieval strategies—reasoning chains are determined by document structure, not prompt framing. Consequently, pairwise judgments find nothing to discriminate, PageRank converges to near-uniformity. \textbf{Across-trial} ($\text{StructU}_{\text{across}} < 0.001$) is near zero because all trials agree on uniformity—genuinely nothing to rank. \textbf{Within-trial} ($\text{StructU}_{\text{within}} \approx \log 5 = 1.609$) reaches theoretical maximum, reflecting flat PageRank where no response dominates.

\paragraph{Contrast with Math-Synth incorrect.} Both have $\text{Self-ConsU} = 0$ (unanimous wrong answer), yet structural profiles diverge sharply. Math-Synth involves \emph{endogenous} error (negation miscounting) where different strategies produce detectably different error paths, yielding preference instability and $\text{StructU}_{\text{across}} = 0.035$. HotpotQA involves \emph{exogenous} failure (missing context) where no reasoning diversity can compensate for absent evidence, producing substantively identical responses and $\text{StructU}_{\text{across}} < 0.001$. StructU distinguishes these unanimous failure regimes—one flagged unreliable (0.035), the other low-uncertainty (<0.001)—but cannot detect failures leaving no trace in preference structure.

\subsubsection{Correct Example ($\text{Self-ConsU} = 0$, $\text{StructU}_{\text{across}} < 0.001$)}

\paragraph{Task.} A HotpotQA question: \emph{What creature of American folklore gained notoriety in 1964?} Retrieved context discusses several folklore creatures (Teakettler, Hidebehind, Chessie) but none mention 1964. All responses correctly identify this gap and abstain.

\paragraph{Observation.} Every response executes identical retrieval: scan all documents for "1964" and folklore creatures → identify Documents 1, 5, 8 as partially relevant (creatures but no 1964) → note closest match is Chessie (1977/1980s sightings) → conclude information absent → abstain. Variation is surface-level only: R1 adds self-check; R2 lists documents; R3 uses Socratic framing; R4 decomposes; R5 casts as date-retrieval pattern.

The uncertainty profile is statistically indistinguishable from the incorrect example: $\text{StructU}_{\text{across}} < 0.001$, $\text{StructU}_{\text{within}} = 1.608$, $\pi_{\max}/\pi_{\min} = 1.08$, all CVs<0.015. Table~\ref{tab:hotpot_correct_pagerank} shows near-uniform PageRank frozen across trials—virtually identical to the incorrect example (Table~\ref{tab:hotpot_incorrect_pagerank}).

\begin{table}[t]
\centering
\small
\begin{tabular}{lccccc}
\toprule
       & R1    & R2    & R3    & R4    & R5    \\
\midrule
Trial 0 & 0.203 & 0.207 & 0.198 & 0.198 & 0.194 \\
Trial 1 & 0.202 & 0.206 & 0.199 & 0.197 & 0.196 \\
Trial 2 & 0.204 & 0.206 & 0.197 & 0.198 & 0.195 \\
Trial 3 & 0.203 & 0.207 & 0.198 & 0.197 & 0.195 \\
Trial 4 & 0.202 & 0.206 & 0.198 & 0.198 & 0.196 \\
\midrule
Mean    & 0.203 & 0.206 & 0.198 & 0.198 & 0.195 \\
CV      & 0.004 & 0.002 & 0.003 & 0.003 & 0.004 \\
\bottomrule
\end{tabular}
\caption{Per-trial PageRank distributions for the HotpotQA correct example. Distributions are near-uniform and frozen across trials, yielding $\text{StructU}_{\text{across}} < 0.001$—statistically indistinguishable from the incorrect example (Table~\ref{tab:hotpot_incorrect_pagerank}).}
\label{tab:hotpot_correct_pagerank}
\end{table}

\paragraph{Significance.} Identical collapse on correct and incorrect examples demonstrates this is a \emph{task structure} property, not error status. Retrieval over fixed documents produces prompt-invariant chains regardless of outcome. Different prompts cannot induce different retrieval strategies—chains are determined by document structure, not prompt framing.

This represents a \textbf{boundary condition} for structural uncertainty. On Math-Synth, StructU successfully separated correct from incorrect unanimous agreement ($0.001$ vs $0.035$, $30\times$ difference) because different prompts induced genuinely different reasoning strategies the judge could differentially rank. On HotpotQA retrieval, StructU produces indistinguishable values (<0.001 both cases) because deterministic retrieval suppresses the reasoning diversity self-preference requires. The preference graph collapses in both cases, rendering StructU structurally uninformative—not because the method is flawed, but because the task affords no structural variation to exploit. This limitation is shared with Self-ConsU, which also reports zero in both cases, highlighting that uncertainty quantification methods relying on response diversity are fundamentally constrained when reasoning is deterministic given input context.

\subsection{Summary and Implications}
\label{app:qual_summary}

The qualitative evidence supports three conclusions:

\paragraph{(1) Structural uncertainty detects errors invisible to self-consistency.} On Math-Synth, the incorrect example exhibits $30\times$ higher across-trial uncertainty than the correct example, despite both having $\text{Self-ConsU} = 0$. The mechanism is that diverse prompt templates elicit structurally distinct reasoning strategies on mathematical tasks, and the model's inability to stably rank these strategies when all are flawed produces the across-trial uncertainty signal.

\paragraph{(2) Preference graph collapse explains the HotpotQA limitation.} On factual retrieval, different prompt templates cannot elicit different reasoning paths because the retrieval process is determined by the document set. The resulting identical reasoning chains produce near-uniform, stable PageRank distributions ($\text{StructU}_{\text{across}} \approx 0$), eliminating the structural signal regardless of correctness.

\paragraph{(3) The collapse signature is itself diagnostic.} Near-zero across-trial uncertainty combined with near-maximum within-trial uncertainty ($\text{StructU}_{\text{across}} \approx 0$, $\text{StructU}_{\text{within}} \approx \log N$) constitutes a detectable signature indicating that the model lacks a coherent internal quality criterion for the task. This signature can inform practitioners about when to rely on structural versus dispersion-based uncertainty methods: when it is detected, self-preference signals are uninformative and alternative estimators should be preferred.


\subsection{Prompt Templates}
\label{app:prompts}

\subsubsection{Response Generation Prompts}
\label{appendix:diverse_prompts}

We employ five distinct prompt templates to induce diverse reasoning patterns across candidate responses.
This diversity is essential for meaningful pairwise comparisons, as it ensures that differences in solution quality reflect substantive reasoning variations rather than superficial stylistic differences.

\begin{respgenprompt}{Prompt 1: Step-by-step with self-check}
Solve this math problem step by step, then double-check your work for any errors.

Q: \{question\}

<initial_solution>
Let me work through this step by step:
[Show complete solution]
</initial_solution>

<self_check>
Now let me review my work for any mistakes:
[Check each step and correct if needed]
</self_check>

<answer>
ONLY include the numerical final answer here WITHOUT units. Do not include any explanation or working in this section, just the number/value.
</answer>

Please strictly follow above format when presenting the answers.
\end{respgenprompt}

\begin{respgenprompt}{Prompt 2: Think-aloud decision process}
Solve this problem while thinking out loud about your decision-making process at each step.

Q: \{question\}

<thinking_process>
I'm reading the problem and thinking... [explain thought process]
Now I need to decide what approach to take... [explain reasoning]
Let me calculate this step... [show work with internal thoughts]
</thinking_process>

<answer>
ONLY include the numerical final answer here WITHOUT units. Do not include any explanation or working in this section, just the number/value.
</answer>

Please strictly follow above format when presenting the answers.
\end{respgenprompt}

\begin{respgenprompt}{Prompt 3: Socratic guiding questions}
Solve this problem by asking yourself guiding questions at each step.

Q: \{question\}

<socratic_dialogue>
What is the problem asking? [Answer]
What information do I have? [List knowns]
What do I need to find? [Identify unknowns]
What method should I use? [Choose approach]
How do I execute this method? [Show work]
Does my answer make sense? [Verify]
</socratic_dialogue>

<answer>
ONLY include the numerical final answer here WITHOUT units. Do not include any explanation or working in this section, just the number/value.
</answer>

Please strictly follow above format when presenting the answers.
\end{respgenprompt}

\begin{respgenprompt}{Prompt 4: Decomposition into sub-problems}
Break this complex problem into smaller, manageable sub-problems and solve each one.

Q: \{question\}

<decomposition>
Sub-problem 1: [Identify and solve]
Sub-problem 2: [Identify and solve]
Sub-problem 3: [Identify and solve]
Combining results: [Show how sub-solutions combine]
</decomposition>

<answer>
ONLY include the numerical final answer here WITHOUT units. Do not include any explanation or working in this section, just the number/value.
</answer>

Please strictly follow above format when presenting the answers.
\end{respgenprompt}

\begin{respgenprompt}{Prompt 5: Analogical reasoning}
Solve this problem by relating it to a simpler, similar problem you know how to solve.

Q: \{question\}

<analogical_reasoning>
This problem is similar to: [Describe simpler analogy]
In the simpler case, I would: [Show simple solution]
Applying the same logic here: [Adapt to current problem]
</analogical_reasoning>

<answer>
ONLY include the numerical final answer here WITHOUT units. Do not include any explanation or working in this section, just the number/value.
</answer>

Note: For the answer section, provide ONLY the numerical result without any units, explanations, or additional text.

Please strictly follow above format when presenting the answers.
\end{respgenprompt}

\subsubsection{Pairwise Comparison Judge Prompt}
\label{appendix:judge_prompt}

The following prompt elicits pairwise preference judgments with confidence scores.
The structured output format (\texttt{<reasoning>}, \texttt{}, \texttt{}) enables consistent parsing across all trials and models.

\begin{judgeprompt}{Pairwise Judge Prompt: Choose the better response with confidence}
I'm going to show you a question and two different responses to it.
Your task is to decide which response better answers the question AND provide a confidence score.

Question: {question}

Response 1:
{response1}

Response 2:
{response2}

Which response better answers the question? Please evaluate them FAIRLY and CONSISTENTLY based on objective criteria, avoiding any subjective bias.

For your evaluation:
- Focus on accuracy, clarity, completeness, and relevance to the question
- Consider evidence and reasoning quality in each response
- Maintain consistent evaluation standards across all comparisons
- Avoid being influenced by response length, style preferences, or personal opinions
- Evaluate the responses as if they were written by the same person
- If the question is ambiguous or unclear, the best response is one that acknowledges this ambiguity and explains different possible interpretations rather than making assumptions

Please structure your response as follows:
Think step by step about the clarity, accuracy, helpfulness, and overall quality of each response.
Compare them thoroughly, analyzing their strengths and weaknesses.

Response 1
OR
Response 2


Do NOT include any reasoning or explanation in the answer section. Only state "Response 1" or "Response 2".


Provide a confidence score (0-100) indicating how confident you are in your judgment.
- 100 means absolute certainty that your chosen response is better
- 50 means both responses are equally good/bad
- 0-49 means minimal preference for your chosen response
- 51-100 means stronger preference for your chosen response

\end{judgeprompt}

\subsubsection{Verbalized Uncertainty Baseline Prompt}
\label{appendix:verbalized_prompt}

For the verbalized uncertainty baseline , we directly elicit the model's self-assessed confidence.
This prompt produces a single response with an explicit confidence score, which we compare against structural and self-consistency baselines.

\begin{verifierprompt}{Verifier Prompt: JSON-only solution check}
CRITICAL INSTRUCTION: You MUST respond with ONLY a JSON object. NO reasoning, NO explanation, NO other text.

Verify if this solution is correct:

PROBLEM: {problem}

SOLUTION: {solution}

Respond with ONLY this JSON (nothing else):
{"verdict": "PASS", "confidence_correct": 0.95}

Replace PASS with FAIL if incorrect, and set confidence 0.0-1.0.
JSON ONLY. NO OTHER TEXT.
\end{verifierprompt}


\end{document}